\documentclass[manuscript,screen]{acmart}
\usepackage{oz}

\makeatletter
\let\ACM@origbaselinestretch\baselinestretch
\makeatother

\usepackage{indentfirst}

\usepackage{array}
\usepackage{multirow}
\usepackage{booktabs} 
\usepackage{graphicx}
\usepackage{adjustbox}
\usepackage{tabularx}
\graphicspath{./figures}
\usepackage{colortbl}
\usepackage{float}
\usepackage{amsmath}

\usepackage{rotating}
\usepackage{caption}
\usepackage{subcaption}

\newcommand{\added}[1]{\textcolor{black}{{#1}}}
\newcommand{\Rev}[1]{\textcolor{black}{{#1}}}

\usepackage[backgroundcolor=white,bordercolor=black,linecolor=black]{todonotes}

\usepackage{multirow}
\usepackage{tcolorbox}

\usepackage[ruled, vlined, linesnumbered]{algorithm2e}
\newcommand{\nonl}{\renewcommand{\nl}{\let\nl\oldnl}}

\SetKwInput{KwInput}{Input}                
\SetKwInput{KwOutput}{Output} 

\SetAlFnt{\small\sffamily}

\SetCommentSty{mycommfont}

\usepackage{amsmath, listings, amsthm, xspace}

\usepackage{hyperref}
\hypersetup{colorlinks,allcolors=black}

\title{DeepGD: A Multi-Objective Black-Box Test Selection Approach for Deep Neural Networks}

\author{Zohreh Aghababaeyan}
\affiliation{
    \institution{University of Ottawa}
    \city{Ottawa}
    \country{Canada}}
\email{Zohreh.a@uottawa.ca}

\author{Manel Abdellatif}
\affiliation{
    \institution{École de Technologie Supérieure }
    \city{Montreal}
    \country{Canada}}
\email{Manel.abdellatif@etsmtl.ca}

\author{Mahboubeh Dadkhah}
\affiliation{
    \institution{University of Ottawa}
    \city{Ottawa}
    \country{Canada}}
\email{mdadkhah@uottawa.ca}

\author{Lionel Briand}
\affiliation{
    \institution{University of Ottawa}
    \city{Ottawa}
    \country{Canada}}
\affiliation{
    \institution{Lero SFI Centre on Software Research and University of Limerick}
    \country{Ireland}
    }
\email{Lbriand@uottawa.ca}

\begin{document}

\begin{abstract}
    Deep neural networks (DNNs) are widely used in various application domains such as image processing, speech recognition, and natural language processing. However, testing DNN models may be challenging due to the complexity and size of their input domain. Particularly, testing DNN models often requires generating or exploring large unlabeled datasets. In practice, DNN test oracles, which identify the correct outputs for inputs, often require expensive manual effort to label test data, possibly involving multiple experts to ensure labeling correctness. In this paper, we propose \textit{DeepGD}, a black-box multi-objective test selection approach for DNN models. It reduces the cost of labeling by prioritizing the selection of test inputs with high fault-revealing power from large unlabeled datasets. \textit{DeepGD} not only selects test inputs with high uncertainty scores to trigger as many mispredicted inputs as possible but also maximizes the probability of revealing distinct faults in the DNN model by selecting diverse mispredicted inputs. The experimental results conducted on four widely used datasets and five DNN models show that in terms of fault-revealing ability: (1) White-box, coverage-based approaches fare poorly, (2) \textit{DeepGD} outperforms existing black-box test selection approaches in terms of fault detection, and (3) \textit{DeepGD} also leads to better guidance for DNN model retraining when using selected inputs to augment the training set. 
\end{abstract}

\keywords{Deep Neural Network, Test Case Selection, DNN Fault Detection, Multi-Objective Optimization, Deep Learning Model Evaluation, Uncertainty Metrics, Diversity, Model Retraining Guidance}

\ccsdesc[500]{Computing methodologies~Machine learning}
\ccsdesc[500]{Software and its engineering~Software verification and validation}
\ccsdesc[300]{Computing methodologies~Computer vision}

\maketitle
\section{Introduction} 
\label{Sec:Introduction}

Deep Neural Networks (DNNs) have become widely used in a variety of application areas, including image processing~\cite{alam2020survey}, medical diagnostics~\cite{litjens2017survey}, and autonomous driving~\cite{grigorescu2020survey}. However, DNNs may produce unexpected or incorrect results that could lead to significant negative consequences or losses. Therefore, effective testing of such models is crucial to ensure their reliability~\cite{tambon2022certify}.
For testing and enhancing the performance of DNN-driven applications, a significant amount of labeled data is required. \Rev{In many real-world scenarios, there are instances where the availability of labeled data for testing is either limited or absent. This situation becomes particularly pertinent when the trained and tested model is deployed in different operational settings and necessitates further evaluation using operational or real data, typically large and unlabeled. The same challenge extends to the testing of third-party pre-trained DNN models for example, where access to internal model information or training data is typically unavailable. In such cases, these models are commonly subjected to further testing using end-users' data, which is, in general, massive and unlabeled.}
However, obtaining the correct output labels for a large amount of unlabeled data, referred to as oracle information, can be a challenging and resource-intensive task ~\cite{chen2020practical}. \Rev{This process often requires extensive manual effort by domain experts and can be a significant challenge when the volume of unlabeled data is large and resources are limited.} 
\Rev{For instance, in many real-world situations, DNN testing is usually bounded by a limited testing budget, especially when testing is performed on costly simulators such as flight simulators or real hardware (e.g., self-driving cars, medical equipment). These execution environments can be expensive and require substantial computing resources, which in turn limits the number of tests that can be performed~\cite{zolfagharian2024smarla}.} 
This makes it more difficult to effectively test and improve the performance of DNN models.
In order to address these challenges and make DNN testing feasible and cost-effective in practice, it is essential to select a small subset of test inputs that possess a high fault-revealing power. This approach reduces the number of inputs required for DNN testing, making it more cost-effective.

Several test selection approaches for DNN models have been proposed in recent years~\cite{pei2017deepxplore, Ma2018DeepGaugeMT, DeepTestRef, kim2019guiding, sun2019structural, gao2022adaptive, ma2021test, hu2022empirical}. Some of them are white-box since they require access to the internals of the DNN models or the training datasets. Similar to code coverage in traditional software systems~\cite{hemmati2015effective}, researchers have proposed several neuron coverage metrics in the context of DNN test selection to prioritize the selection of test inputs with high coverage scores~\cite{ pei2017deepxplore, Ma2018DeepGaugeMT, DeepTestRef, kim2019guiding, sun2019structural}.
Studies have shown the effectiveness of these approaches for distinguishing adversarial inputs but failed to demonstrate a positive correlation between coverage and DNN mispredictions~\cite{chen2020deep,aghababaeyan2021black, li2019structural, yang2022revisiting}. Furthermore, for some coverage metrics, reaching maximum coverage can be easily achieved by selecting a few test inputs~\cite{ sun2018testing, chen2020deep}, thus putting into question the usefulness of such white-box DNN test selection approaches. Last, the use of white-box DNN test selection approaches is hindered by the requirement for access to the internal structure of the DNN model or the training dataset. This can be a significant limitation, particularly when the model is proprietary or provided by a third party~\cite{aghababaeyan2021black}. 

To alleviate this latter problem, black-box test selection approaches have been proposed, which do not require access to the internal information of the DNN model under test or its training set. 
They generally rely on DNN output uncertainty metrics to guide the selection of test inputs. It has been observed that a test input is likely to be mispredicted by a DNN if the model is uncertain about its output classification, when similar probabilities are predicted for each class ~\cite{feng2020deepgini}. An empirical experiment conducted by Ma \textit{et al.}~\cite{ma2021test} revealed that these metrics have medium to strong correlations with mispredicted inputs, while coverage-based metrics have weak or no correlation. One issue though is that if the model is not sufficiently trained, outputs cannot be trusted, thus leading to unreliable uncertainty results and test case selection~\cite{li2019boosting}.

Another black-box strategy that has been successfully applied in testing traditional software systems is to maximize diversity among test cases~\cite{biagiola2019diversity, hemmati2015prioritizing, feldt2016test}.
However, only a few studies have investigated its effectiveness in testing DNN models~\cite{zhao2022can,gao2022adaptive,aghababaeyan2021black}. Zhao \textit{et al.}~\cite{zhao2022can} studied state-of-the-art (SOTA) test selection approaches for DNNs and found that they do not guarantee the selection of diverse test input sets. Furthermore, in our prior work~\cite{aghababaeyan2021black}, we compared the fault-revealing power of several diversity metrics with coverage criteria. We found that geometric diversity (GD)~\cite{ kulesza2012determinantal} outperforms coverage criteria in terms of fault-revealing power and computational time. However, we did not investigate how such a metric can be used for DNN test selection.

In this paper, we propose \textit{DeepGD} \Rev{(Deep Geometric Diversity)}, a black-box multi-objective search-based test selection approach. It relies on the Non-dominated Sorting Genetic Algorithm (NSGA-II)~\cite{deb2002fast} to select test inputs that aim at revealing a maximal number of diverse faults in DNN models within a limited testing budget. The latter is mostly determined by test input labeling effort and, for larger models, their execution time. \textit{DeepGD} can therefore be used to select a subset of test inputs to label from a large unlabeled dataset. 
The strategy of \textit{DeepGD} is twofold: (1) trigger as many mispredicted inputs as possible by selecting inputs with high uncertainty scores, and (2) maximize the probability of revealing distinct faults in the DNN model by selecting diverse mispredicted inputs.

Similar to failures in traditional software systems, many mispredictions result from the same faults (causes) in the deep learning model and are therefore redundant~\cite{aghababaeyan2021black,fahmy2021supporting, zolfagharian2023search}.
Misprediction rates\footnote{Misprediction rate = 1 - Accuracy} can therefore be misleading when assessing a test selection approach, if a large number of mispredictions stem from the same fault. This is also why, for similar reasons, traditional software testing studies typically focus on faults, not failures~\cite{pachouly2022systematic,hall2011systematic, pan2023atm, pan2022test}.
Therefore, to validate the effectiveness of \textit{DeepGD}, we use a fault estimation approach proposed in~\cite{aghababaeyan2021black}, which is based on clustering mispredicted inputs based on their features. Each cluster represents a distinct fault since it contains similar inputs that are mispredicted due to the same root causes. 
More specifically, we present an empirical evaluation of the effectiveness of  \textit{DeepGD} for selecting test input sets with high fault-revealing power. Since our fault estimation approach is tailored to image datasets and relies on clustering mispredicted inputs according to image features, we conduct our experiments on image classification models. They include five widely-used DNN models and four image recognition datasets. The results are compared with nine SOTA baseline selection approaches, both white-box and black-box, that have been recently published~\cite{pei2017deepxplore,kim2019guiding,gao2022adaptive,ma2021test,feng2020deepgini,Ma2018DeepGaugeMT}.

The experimental results demonstrate that \textit{DeepGD} yields a statistically significant and consistent improvement compared to SOTA approaches in terms of its ability to reveal DNNs faults. 
Specifically, results show that \textit{DeepGD} is consistently the best approach, up to 4 percentage points (pp) better than the second-best alternative and 14 pp better than the worst black-box alternative, when excluding random selection (RS) since RS showed far inferior results in general. It is important to note that the ranking of alternatives varies across datasets, models, and test set sizes. Consequently, selecting any approach other than DeepGD may end up being the worst choice and lead to significant differences in performance. Further, we also investigate the effectiveness of \textit{DeepGD} in guiding the retraining of DNNs and demonstrate that it consistently provides better results with that respect as well.  
\textit{DeepGD} is therefore the only technique we can confidently recommend.

To summarize, the key contributions of this paper are as follows:
\begin{itemize}

    \item We propose a black-box test selection approach (\textit{DeepGD}) for DNNs that relies on a customized multi-objective genetic search and uses both diversity and uncertainty scores to guide the search toward finding test input sets with high fault-revealing power. 
    
    \item Unlike existing test selection approaches, we consider in our validation a clustering-based approach to estimate faults in DNN models since test input sets typically contain many similar mispredicted inputs caused by the same problems (faults) in the model~\cite{aghababaeyan2021black,fahmy2021supporting}. We thus obtain more meaningful evaluation results, similar to common practice in testing research. 
     
    \item We conduct a large-scale empirical evaluation to validate \textit{DeepGD} by considering five DNN models, four different datasets, and nine SOTA test selection approaches for DNNs as baselines of comparison. We show that \textit{DeepGD} provides better guidance than baselines for (1) selecting inputs with high fault-revealing power, and (2) improving the performance of the model through retraining based on an augmented training set. 

    \item \Rev{We study the diversity of inputs selected by \textit{DeepGD} and SOTA baselines to assess whether our approach not only leads to test inputs with higher fault-revealing capabilities but also promotes greater diversity in the selected test input sets. We thus confirm that \textit{DeepGD} indeed selects more diverse input sets compared to baselines.}
    \item \Rev{We analyze the execution time of \textit{DeepGD} and other baselines and show that they are not computationally expensive. Though \textit{DeepGD} exhibits longer execution times, these remain practical and will have little impact in practice. \textit{DeepGD}'s improved fault detection and retraining guidance more than compensate for them.}

\end{itemize}

The remainder of the paper is structured as follows. Section~\ref{Sec:Approach} presents our test selection approach. Section~\ref{Sec:Evaluation} presents our empirical evaluation. Section~\ref{Sec:Results} outlines our results. Section~\ref{sec:Threats} describes the threats to the validity of our study. Sections~\ref{Sec:RW} and \ref{Sec:Conclusion} contrast our work with related work and conclude the paper, respectively.

\section{Approach: Reformulation as an NSGA-II Search Problem }
\label{Sec:Approach}

A central problem in testing DNNs, especially when the labeling of test data is costly, is the selection of a small set of test inputs with high fault revealing power. In this paper, we aim to support the testing of DNN models by relying on \textit{DeepGD}, a black-box search-based test selection approach using a genetic algorithm to select small sets of test inputs with high fault-revealing power in DNN models. Intuitively, testers should select a set of diverse test inputs with high failure probabilities in order to be more likely to detect as many diverse faults as possible~\cite{zohdinasab2021deephyperion}. Due to the combined high labeling cost of test inputs and large input space, we rely on NSGA-II to select test inputs with high fault-revealing capability. Such inputs are then selected for labeling and will be used to effectively test the DNN model. 
We choose NSGA-II since it is widely used in the literature and showed its performance to solve many search-based test selection problems~\cite{hemmati2013achieving,wang2015cost}. We also rely on NSGA-II since it is specifically adapted to our multi-objective search problem. It tries to find solutions with diverse trade-offs between fitness functions instead of covering all fitness functions separately (which is the case of MOSA, the Many Objective Sorting Algorithm ~\cite{MOSA7102604} for example). 
Specifically, the search is driven by two objectives: (1) maximizing the uncertainty score of the test inputs to trigger a maximum number of mispredictions, and (2) maximizing the diversity of the test input set to trigger diverse mispredictions caused by distinct faults. 
To properly translate the process into a search problem using NSGA-II, we need to define the following elements.

\subsection{Individuals and Initial Population}

In genetic algorithms, individuals consist of a set of elements called genes. These genes are connected and form an individual that is also called a solution. In our approach, we assign a unique identifier $id_i$ to each input $i$ in the test dataset where $id_{1 \leq i \leq n} \in [1,2,..,n]$ and $n$ is the size of the test dataset. Our test selection problem has a fixed testing budget $\beta<n$ which corresponds to the total number of inputs selected from the original test dataset to test the DNN model. In our search problem, an individual corresponds to a subset of inputs of size $\beta$. Each gene forming an individual corresponds to an $id$ of a test input in the test dataset. 
In our context, each individual contains $\beta$ distinct test inputs. We use random selection to build our initial population of individuals.


\subsection{Fitness Functions} \label{subSec:Fitness}

The main objective of our approach is to select unlabeled inputs that possess a strong ability to reveal diverse faults.
These faults manifest themselves as diverse mispredicted inputs. To generate the latter, our search process is guided by two fitness functions. The first function aims to maximize the uncertainty score of the selected inputs, as it is moderately to strongly correlated with mispredictions~\cite{ma2021test,feng2020deepgini,weiss2022simple, gao2022adaptive}. By maximizing uncertainty scores, we are thus more likely to trigger a larger number of mispredicted inputs~\cite{ma2021test}. The second function aims to maximize the diversity among the selected inputs. This approach increases the probability of identifying distinct faults. \Rev{By simultaneously maximizing these two fitness functions, we aim to select a diverse set of mispredicted inputs that are due to distinct faults, thereby maximizing the fault-revealing power of the selected test input set.}
We will describe next the two fitness functions and detail how to compute them. 

\subsubsection{\textbf{Gini Score}} \label{SubSec:Ginimethod}
We consider the \textit{Gini} score to estimate the likelihood of a test input being mispredicted. Feng \textit{et al.}~\cite{feng2020deepgini} proposed this metric to measure the classification uncertainty of DNN models and therefore identify potential failing test inputs. Intuitively, a test input is likely to be misclassified by a DNN if the model is uncertain about the classification and outputs similar probabilities for each class~\cite{feng2020deepgini}.
We choose this metric since it has been widely used in the literature, showed good performance in prioritizing test inputs for DNN models, and has a medium to strong correlation to misprediction rates~\cite{ma2021test,feng2020deepgini,weiss2022simple, gao2022adaptive}. It is also a black-box metric that only requires the output probabilities of DNN models as we will describe in the following.
Given a test input $x$ and a DNN model that outputs $DNN(x)=<P_{x_{1}}, P_{x_{2}},...,P_{x_{m}}>$, where $P_{x_{i\in [1,..,m]}}$ is the probability that input $x$ belongs to class $C_i$ and $m$ is the total number of classes, the \textit{Gini} score of the test input $x$ is defined as:
\begin{equation}\label{Eq:GiniScore}
 \xi(x)= 1- \sum_{i=1}^{m} P_{x_{i}}^{2}
\end{equation}

The higher the \textit{Gini} score, the higher the DNN's uncertainty. We compute the \textit{Gini} score of a subset $S=\{s_1, s_2, ...., s_\beta\}$ of size $\beta$ by computing the average \textit{Gini} score of all inputs in the subset as follow: 
\begin{equation}\label{Eq:GiniScore2}
 Gini(S)= \frac{\sum_{i=1}^{\beta} \xi (s_{i}) }{\beta} 
\end{equation}

\subsubsection{\textbf{Geometric Diversity}}

We consider geometric diversity, one of the widely used metrics to measure the diversity of test input sets ~\cite{aghababaeyan2021black, kulesza2012determinantal, gong2014diverse}. In a previous study~\cite{aghababaeyan2021black}, several coverage and diversity metrics were investigated and results showed that the GD metric is positively correlated to DNN faults and outperforms SOTA coverage metrics in terms of fault-revealing capabilities. In other words, when the geometric diversity of a test input set increases, its fault-revealing power increases as well since the diverse test input set will cover more faults in the DNN model. Furthermore, GD is a black-box diversity metric that requires neither knowledge about the model under test nor access to the training set. Consequently, we have relied on this metric as fitness function to guide the search towards finding a diverse test input set with high fault-revealing capability. \\

\textbf{Feature Extraction.} In order for diversity to account for the content of images, we need to first extract features from each input image and then compute the diversity based on those extracted features. 
\Rev{In our work, we use VGG-16~\cite{simonyan2014very} which is widely recognized as one of the SOTA models for image feature extraction. It is a pre-trained convolutional neural network model which was already trained on ImageNet~\cite{deng2009imagenet}, an extensive dataset including over 14 million labeled images. As a result, it has learned rich feature representations for a wide range of images and datasets~\cite{simonyan2014very}. VGG-16 has been extensively used in various image recognition problems and has been reported to be highly accurate~\cite{aghababaeyan2021black, sharma2022deep, mousser2019deep, kaur2019automated}. Despite its simple architecture, the VGG16 model has shown comparable performance to more complex feature extraction techniques such as ResNet50 and VGG19~\cite{mousser2019deep}. We therefore rely on this feature extraction technique to produce high-quality feature representations, thereby enhancing the accuracy of our diversity measures.}
 
We extract features of each image in the test input set $S$ using VGG-16 and generate the corresponding feature matrix \textbf{$F = (f_{ij})\in R^{n*m}$} where $n$ is the number of input images in  $S$, and $m$ is the number of features. Each row of this matrix represents the feature vector of an image, and each column ($F_j$) represents a feature. 
\Rev{It is worth mentioning that, in order to ensure compatibility of certain image datasets with VGG-16, specific image pre-processing techniques are applied before feature extraction. These techniques involve operations such as image resizing and color channel duplication.}
After extracting the feature matrix, we normalize it as a pre-processing step to eliminate the dominance effect of features with large value ranges and to make the computation of the selected diversity metrics more scalable~\cite{aghababaeyan2021black}. We apply the \textit{Min-Max normalization} per feature, and transform the maximum and minimum values of that feature to one and zero, respectively. For every feature $F_j$ in the feature matrix \textbf{$F$} where $F_j(i)$ is the value of feature number $j$ for $i^{th}$ input image in $S$, the normalized feature $F'_j$ is calculated as follows: 
\begin{equation}\label{Eq:MinMaxNormalization}
 F'_j(i) = \frac{ F_j(i) - min(F_j)}{max(F_j)-min(F_j)}
\end{equation}

\begin{figure}[ht]
\includegraphics[width = \columnwidth]{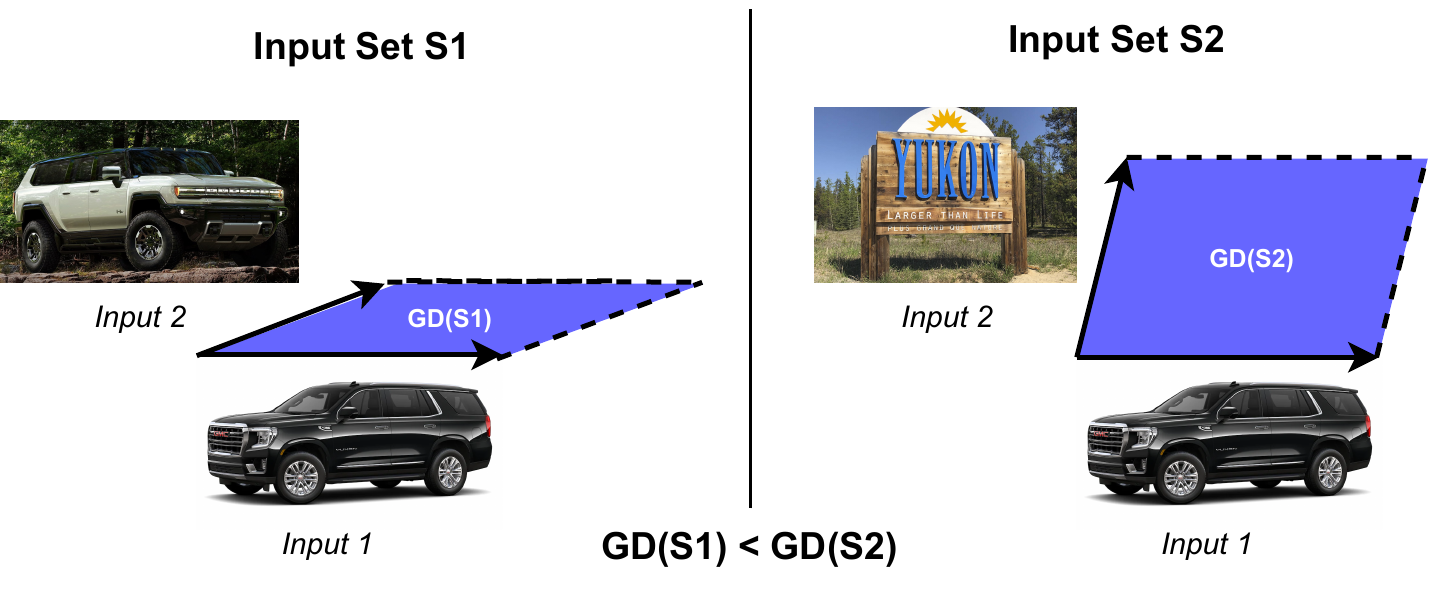}

\centering
\caption{\Rev{Illustration of the geometric diversity metric~\cite{aghababaeyan2021black}}}
\label{fig:ExampleGeometricDiversity}
\end{figure}

\begin{figure}[ht]
\includegraphics[width = 10cm]{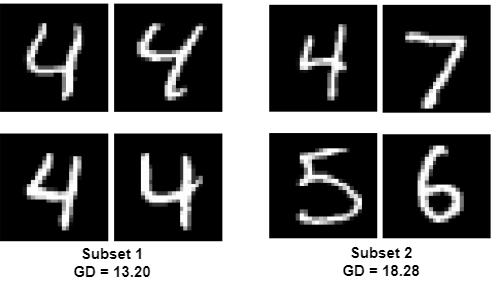}

\centering
\caption{Examples of geometric diversity scores for two different subsets}
\label{fig:example}
\end{figure}

\textbf{Computation of Geometric Diversity Scores.} After extracting the feature vectors, we calculate the geometric diversity of the test inputs as our second fitness function with the goal of selecting as many diverse test input sets as possible.
Given a dataset $\mathbb{D}$, the normalized feature matrix \textbf{$F'$} where $F'_S$ represents feature vectors of a subset $S {\subseteq} \mathbb{D} $, the geometric diversity of $S$ is defined as: 
\begin{equation}\label{Eq:GeometricDiversity}
 GD(S) = det(F^\prime_S * {F^\prime_S}^{T} )  
\end{equation}
which corresponds to the squared volume of the parallelepiped spanned by the rows of $F^\prime_S$, since they correspond to vectors in the feature space. The larger the volume, the more diverse is $S$ in the feature space. We illustrate in Figure~\ref{fig:ExampleGeometricDiversity} an example of the geometric diversity of two input sets, \Rev{\textit{S1} and \textit{S2}, including two inputs each. For the sake of simplicity, each input in this example is represented by its corresponding two-dimensional feature vector.} The geometric diversity of an input set corresponds, to the squared volume of the parallelepiped spanned by the feature vectors of the input set, which is here a simple plane highlighted in blue \Rev{(i.e., GD(S1) and GD(S2))}. The larger the volume \Rev{(i.e., the geometric diversity score)}, the more diverse the input set, as we can see in Figure~\ref{fig:ExampleGeometricDiversity}.
Moreover, Figure~\ref{fig:example} includes two subsets of the MNIST dataset~\cite{deng2012mnist}, each containing four images, evaluated with their respective GD scores. Subset 1, a low-diversity test set, includes images with greater similarity, reflected by its GD score of 13.20. In contrast, Subset 2 showcases a high-diversity test set, characterized by a wider range of image variations, as indicated by its higher GD score of 18.28.

\subsubsection{\textbf{Multi-Objective Search}}
 
We need to maximize in our search the two aforementioned fitness functions to increase the fault-revealing capability of the selected test input set. This is therefore a multi-objective search problem that can be formalized as follows: 

\begin{equation}
\label{eq:fitness}
\begin{split}
    \max\limits_{S \subset \mathbb{D}} \:\:& Fitness(S) =  (\textit{Gini}(S),GD(S))\\ 
\end{split}
\end{equation}
where $\mathbb{D}$ is the test dataset, $S$ is a subset of test inputs of size $\beta$, function $Fitness: S \xrightarrow{} \mathbf{R}^2$ consists of two real-value objective functions $(Gini(S),GD(S))$, and $\mathbf{R}^2$ is the objective space of our optimization problem. \Rev{Given our two objectives, the “\textit{max}” operator in Eq.~\ref{eq:fitness} returns non-dominated solutions forming a Pareto front in the space defined by Gini and GD~\cite{deb2002fast}, i.e., solutions where at least one of the objective function values is the maximum and the other remains unsurpassed by other solutions.}

\subsection{Genetic Operators}~\label{Subsec:Operators}

We describe below the two genetic operators in our test selection approach. The first operator is crossover, which generates new offspring by cutting and joining high-fitness parents. The second operator is mutation which introduces small changes in individuals by mutating specific genes to (1) make the search more exploratory and (2) thus attempt to increase the uncertainty score and the diversity among individuals in the population. We provide below a detailed description of how we customized these genetic operators to fit our test selection problem.

\subsubsection{\textbf{Crossover}}

The crossover operator takes as input two parents (i.e., two subsets) and generates new offspring by slicing and joining the different parts of the selected parents. Let $S_1$ and $S_2$ be the two selected parents for crossover. We provide next an example of selected parents: 

\begin{equation*}
\small
\begin{split}
S_1 = \{Input_1, Input_2, Input_3, \cdots, Input^{s_1}_i,\cdots,   Input_4, Input_5\} 
\end{split} 
\end{equation*}
\begin{equation*}
\small
\begin{split}
S_2 = \{Input_6, Input_7, Input_8, \cdots, Input^{s_2}_i, \cdots, Input_9, Input_{10}\} 
\end{split} 
\end{equation*}
\Rev{where (1) $Input^{s_1}_i$ and $Input^{s_2}_i$ denote the $i^{th}$ input in $S_1$ and $S_2$, respectively and (2) $Input_{1\leq j \leq 10}$ denote all inputs in each individual.} 
Before applying the crossover operator, we start by sorting the inputs forming each parent according to their \textit{Gini} scores. Inputs with higher \textit{Gini} scores are placed at the beginning of each corresponding parent, as in the example below. 

\begin{equation*}
\small
\begin{split}
S_1 = \{Input_3, Input_1, Input_5,\cdots, Input^{s_1}_i,\cdots, Input_4, Input_2\} 
\end{split} 
\end{equation*}
\begin{equation*}
\small
\begin{split}
S_2 = \{Input_7, Input_{10}, Input_6, \cdots, Input^{s_2}_i,\cdots, Input_9, Input_8\} 
\end{split} 
\end{equation*}

Such reordering will help with the creation of potential high-fitness offspring with high uncertainty scores as we will explain in the following. After such sorting, we randomly select a crossover point using the uniform distribution. To form the first offspring, we slice and join the first parts of each parent based on the crossover point. Such offspring includes inputs with the highest \textit{Gini} scores thanks to sorting. Finally, the second offspring is formed by joining the remaining parts of the selected parents. Since this offspring has inputs with the lowest \textit{Gini} scores from both parents, it will be potentially discarded later by the selection operator or improved by the mutation operator as we will detail in the next section. Assuming that the crossover point is at position $i$ in the example above, the generated offspring would be: 

\begin{equation*}
\small
\begin{split}
Offspring_1 = \{Input_3, Input_1, \cdots,Input^{s_1}_{i}, Input_7, Input_{10}, \cdots, Input^{s_2}_{n-i}\} 
\end{split} 
\end{equation*}
\begin{equation*}
\small
\begin{split}
Offspring_2 = \{Input^{s_1}_{i+1}, \cdots, Input_4, Input_2, Input^{s_2}_{n-i+1}, \cdots, Input_9, Input_8\} 
\end{split} 
\end{equation*}

However, applying the crossover on the selected parents may lead to redundant inputs in the created offspring. Since one of our search goals is to maximize the diversity of the selected subsets, we remove such redundant inputs (and therefore increase the diversity of the offspring) by replacing them with random inputs that are not present in the created offspring. We should note that we do not use the GD metric to customize the crossover operator since it is computationally expensive. Instead, as described in the next section, we employ it in the mutation operator since mutations occur less frequently.

\subsubsection{\textbf{Mutation}} \label{subsec:mutation}

After completing the crossover, the offspring are selected for mutation to randomly change some genes that have (1) a low $Gini$ score, and (2) a low contribution to the diversity of the selected offspring. Therefore, the mutation operator is considered in our search approach as a corrective operator with the goal of improving the offspring in terms of both fitness functions. As an example, let us assume that $S_m$ is the test input set to mutate. We select 2\% of the inputs from $S_m$ that have the lowest $Gini$ score. 
We mutate half of these inputs that have the least contribution to the diversity of $S_m$. Consequently, only 1\% of its genes are mutated. \Rev{This choice was inspired by recommendations for multi-point mutations in the literature~\cite{yuchi2019bi,ma2023improved,verma2021comprehensive}. We should recall that to achieve our desired final mutation rate of 1\%, we initially select 2\% of the genes. We compute the contribution to the diversity of the latter and retain the 1\% of the genes that show the highest similarity. This allows us to contain the high cost of computing diversity associated with all genes while focusing mutation on genes contributing the least to diversity.} 




To measure the contribution of an $input_i$ to increasing the diversity in $S_m$, we measure the difference $GD_{diff}(S_m, i)$ between $GD(S_m)$ and $GD(S_m\setminus\{input_i\})$. 
The lower the difference, the more similar the input compared to the other ones in $S_m$. Inputs with low differences should therefore be mutated to accelerate our search process. 

An example of offspring generated through mutation is provided below. $S_m$ is the original offspring and $S_m'$ is the mutated one. Suppose that $S_m$ has 10 genes. We mutate $Input_{1}$ since it has a low $Gini$ score and a low contribution to the diversity of the selected offspring. \Rev{We mutate this gene by replacing it with a randomly selected input from the population. In this example, we replace $Input_{1}$ with $Input_{13}$. To maintain diversity, we make sure that the randomly selected input is distinct from any existing input in the original offspring.}

\begin{equation*}
\small
\begin{matrix}
S_m: & Input_3 & \mathbf{Input_{1}} & Input_5 & \cdots & Input_7 & Input_{10}  \\
$\textit{Gini Score:}$  & \textbf{70\%} & \textbf{71\%}  & 75\%  &\cdots &  95\% & 96\%   \\
\textit{$GD_{diff}$:}  & 60 & \textbf{10 } &  \cdots &  \\
\end{matrix}
\end{equation*}
\begin{equation*}
\small
\begin{matrix}
&S_m': && Input_3 & \mathbf{Input_{13}} & Input_5 & \cdots & Input_7 & Input_{10}
\end{matrix}
\end{equation*}

It should be noted that each genetic operator has a distinct emphasis on a particular fitness function. The crossover operator primarily aims to generate offspring with high uncertainty scores, while the mutation operator also helps improving the offspring in terms of diversity. Crossover and mutation operators therefore complement each other.


\subsection{Search Algorithm}

\begin{algorithm}[t!]

\DontPrintSemicolon
  \KwInput{The initial population ($P$) that is a set of test subsets, the number of generations ($G$), the crossover rate ($CrossRate$), the mutation rate ($MutationRate$)  }

  \KwOutput{A set of individuals ($\alpha$) maximizing the fitness function ($Fitness$)}
    $gen \xleftarrow{} 0$ \;
    \While{$  gen \leq G $}{
    $P_{new} \xleftarrow{} \emptyset$\;
    $P_{Cross} \xleftarrow{} \emptyset$\;
    $P_{Cross} \xleftarrow{} Cross(P, CrossRate)$\;
    $P_{new} \xleftarrow{} Mutate(P_{Cross}, MutationRate)$\;
    $P \xleftarrow{}  Select (P, P_{new}, Fitness)$ \;
    $Update$($\alpha$) \;
    $gen \xleftarrow{} gen +1$ \;
   
    }
    \KwRet{$\alpha$}\;
    
\caption{High-level NSGA-II algorithm}

\label{Algorithm-overview}
\end{algorithm}

As explained earlier, the main objective of our search algorithm is to select from a large unlabeled test dataset, an input set of size $\beta$ with a high fault-revealing power in order to (1) reduce labeling cost, and (2) effectively test the DNN model. Algorithm~\ref{Algorithm-overview} describes at a high level our NSGA-II search process. Assuming $P$ is the initial population, that is a set of $|P|$ randomly selected test input sets, the algorithm starts an iterative process, taking the following actions at each generation until the maximum number of generations ($G$) is reached (lines 2-10). The search process includes the following steps. First, we create a new empty population $P_{new}$ (line 3). Second, based on predefined crossover and mutation rates, we create offspring using crossover ($Cross$) and mutation ($Mutate$) operators and add newly created individuals to $P_{new}$ (lines 4-6). Third, we calculate the fitness of the new population by computing \textit{Gini} and geometric diversity scores of each subset in the population and select individuals for survival using tournament~\cite{Miller96} selection ($Select$) (line 7). Finally, we update the archive $\alpha$ (line 8) by storing the fittest individuals in the population then we move to the next generation (line 9). The non-dominated solutions contained in the population of the last generation $\alpha$ represent the final \textit{Pareto} front of the genetic search. \Rev{To select the final solution from the Pareto front, we use the knee point technique~\cite{branke2004finding} as recommended by several existing studies~\cite{messaoudi2018search}. The solution that corresponds to the knee point has been shown to provide the best trade-off between the different objectives in many multi-objective problems~\cite{messaoudi2018search,branke2004finding}. To calculate the knee point, we first identify the \textit{ideal point}, where the coordinates represent the highest fitness scores (\textit{$GD_{max}$},\textit{$Gini_{max}$}) obtained from all solutions on the Pareto front, with each fitness function evaluated independently. Given the solutions at the Pareto front $P= \{S_1,...,S_{ p}\}$, the knee point $S_k \in P$ is the solution that minimizes
the distance $\sqrt{(GD_{max} \text{--} GD(S_i))^2 + (Gini_{max} \text{--} Gini(S_i))^2}$, for all $S_i \in P$.} 

We set the NSGA-II parameters in our search algorithm as follows. The crossover rate $CrossRate$ is equal to 75\%, given recommended values between 45\% and 95\%~\cite{briand2006using,cobb1993genetic, messaoudi2018search}. The mutation rate $MutationRate$ is equal to 70\%. Although the recommended mutation rate in the literature is proportional to the length of the individual~\cite{briand2006using, cobb1993genetic, messaoudi2018search}, we have not used such a rate since we used mutation for a different purpose than just exploration, i.e., to help increase both uncertainty and diversity scores. According to our preliminary experiments (which we do not include in this paper), we found that smaller mutation rates, consistent with literature recommendations, led to poorer search results. The population size is set to 700 individuals. Furthermore, the stopping criterion is set to 300  generations. We should note that the number of generations was determined through empirical evaluation. This was done by monitoring the evolution of fitness functions over multiple generations using various datasets and models. The maximum number of generations was carefully selected to ensure the convergence of the search process.  
Finally, we used Google Colab and the Pymoo library~\cite{blank2020pymoo} to implement our genetic search.

\section{Empirical Evaluation}
\label{Sec:Evaluation}

This section describes the empirical evaluation of \textit{DeepGD}, including the research questions we address, the datasets and DNN models on which we perform our assessments, our experiments, and our results. 

\subsection{Research Questions} \label{Sec:RQs}

Our empirical evaluation is designed to answer the following research questions. \\

\noindent \textbf{RQ1. Do we find more faults than existing test selection approaches with the same testing budget?}
 Similar to traditional software testing, selecting a subset of test inputs with high fault-revealing ability is highly important in DNN testing, as it should increase the effectiveness of the testing process while reducing input labeling costs. Consequently, we aim in this research question to compare the effectiveness of our approach (\textit{DeepGD}) with existing baselines in terms of their ability to reveal faults in DNNs, while considering the same testing budget. Identifying the source of failures in traditional software systems is relatively straightforward due to the clear and explicit decision logic in the code. However, in DNNs, this task is more challenging as the complexity and non-linearity of the decision-making process make it difficult to determine the cause of the failure. Hence, many papers rely on mispredictions~\cite{feng2020deepgini,pei2017deepxplore,kim2019guiding} for test selection evaluation.
However, similar to failures in traditional software systems, many mispredicted inputs can be due to the same faults in the DNN model and are therefore redundant~\cite{fahmy2021supporting,attaoui2022black}. When selecting inputs on a limited budget, we should therefore avoid similar or redundant mispredictions as they do not help reveal additional root causes or faults in DNN models. 
To accurately answer this research question, we thus rely on a clustering-based fault estimation approach~\cite{aghababaeyan2021black} that we describe in section~\ref{Sec:Faults}, to investigate the effectiveness of the test selection approaches in detecting faults for a fixed testing budget. \\

\noindent \textbf{RQ2. Do we more effectively guide the retraining of DNN models with our selected inputs than with baselines?} 
Retraining DNN models with unseen inputs carefully selected based on DNN testing results is expected to enhance the model's performance compared to that of the original training set. More specifically, it is highly recommended to retrain DNNs with inputs that have the potential to lead to mispredictions~\cite{attaoui2022black,gao2022adaptive,feng2020deepgini}. Consequently, we aim in this research question to investigate the effectiveness of \textit{DeepGD} in guiding the retraining of DNN models by selecting inputs that will be labeled and used to improve the model through retraining. We will not only measure the accuracy improvement resulting from retraining but also analyze it considering the maximum improvement possible based on the available data for retraining.  \\

\noindent\Rev{\textbf{RQ3. Can \textit{DeepGD} select more diverse test input sets?} 
We aim to assess whether \textit{DeepGD} selects more diverse test input sets compared to test selection baselines. Diversity in test input selection is important since it is correlated to faults as we reported in a prior study~\cite{aghababaeyan2021black}. Specifically, more diverse test input sets reveal more faults in DNN models.
Moreover, selecting diverse test inputs mitigates the risk of bias by ensuring that the testing process is not disproportionately skewed towards specific test cases or testing scenarios. It also prevents the detection of redundant faults in DNN models and thus helps optimize the testing budget and effort, since we are more likely to discover unique faults rather than repeatedly identify the same problems in the DNN model under test. } \\

\noindent\Rev{\textbf{RQ4. How do \textit{DeepGD} and baseline approaches compare in terms of computation time?}
We aim to compare the computation times of \textit{DeepGD} and baseline approaches. Specifically, we aim to examine how the computational times change as the size of the test input sets increases. It is essential to understand this scalability factor, as excessively long computation times could potentially limit the practicality and real-world applicability of the studied test selection approaches.}

\subsection{Faults Estimation in DNNs}\label{Sec:Faults}

Before addressing our research questions, we will touch upon the issue of counting faults in DNN models. The motivation for counting faults, when evaluating selection techniques, stems from our objective of detecting different root causes for mispredictions. Considering misprediction rates to compare test selection techniques is highly misleading as many test inputs can be mispredicted for the same reasons~\cite{aghababaeyan2021black,fahmy2021supporting}. 
Similarly, in traditional software testing, the focus of testers is not on selecting test inputs that maximize the failure rate, which is equivalent to the misprediction rate in our context. The focus is instead on maximizing the number of distinct faults detected in the system. 
According to Chan \textit{et al.}~\cite{chen2010adaptive}, in traditional software testing, failure-causing inputs tend to be dense and close together. The same insight applies to DNN model testing, since similar mispredicted inputs tend to be due to the same fault~\cite{attaoui2022black,aghababaeyan2021black}.

\begin{figure}[h]
\centering
\includegraphics[width=\textwidth]{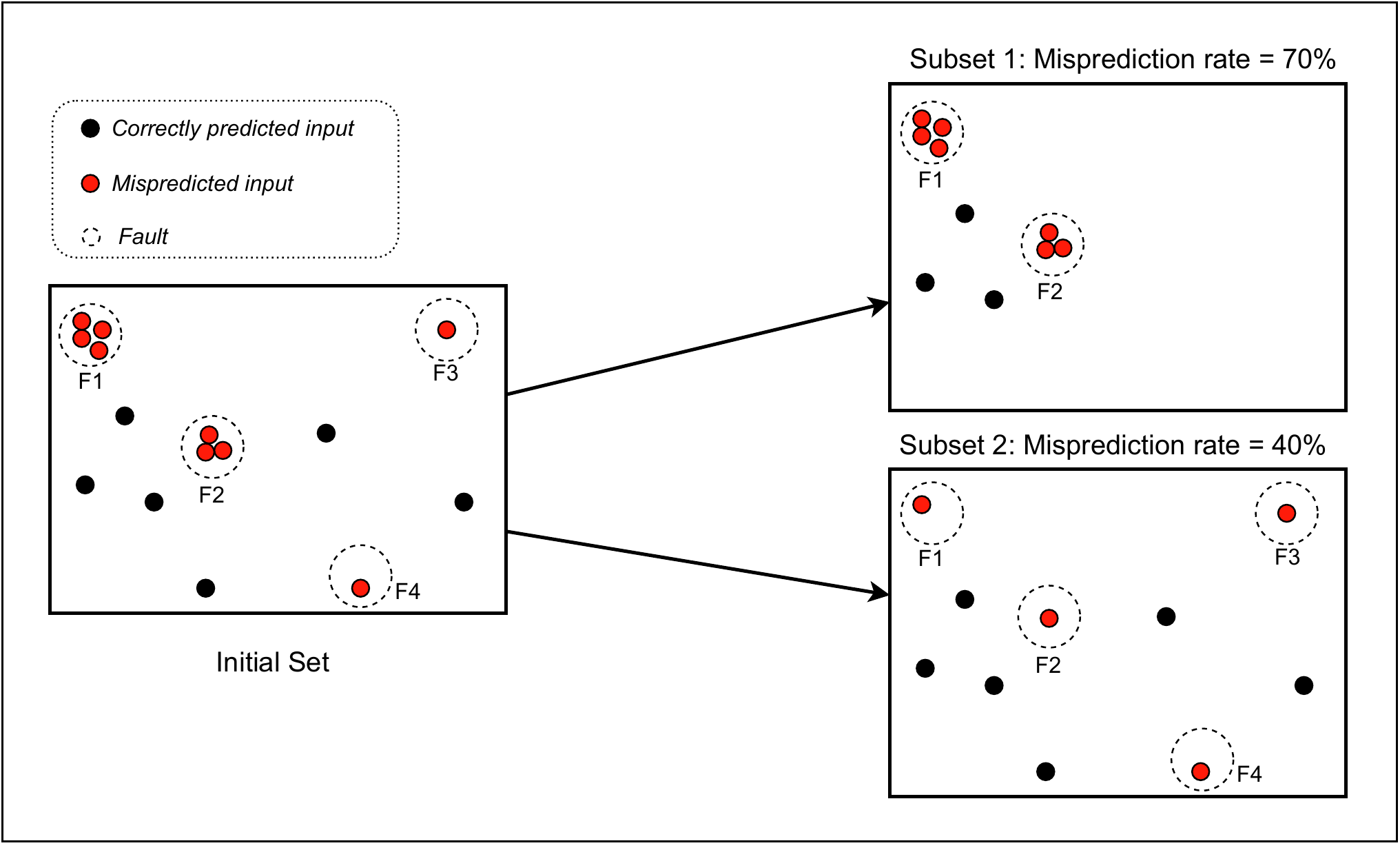}
\caption{Relying on misprediction rates is misleading~\cite{aghababaeyan2021black}}
\label{fig:FaultsFig}
\end{figure}

The above principles should not be different when testing DNNs, where we aim to identify distinct causes of mispredictions (i.e., faults)~\cite{aghababaeyan2021black} and address them, for example, through re-training. We illustrate this point in Figure~\ref{fig:FaultsFig} where a test input set is represented in two-dimensional space. Red and black dots correspond to mispredicted and correctly predicted inputs, respectively. 
We then select two subsets of the same size from the original test input set and measure their misprediction rate (i.e., the number of mispredicted inputs over the size of the subset). As depicted in Figure~\ref{fig:FaultsFig}, subset 1 exhibits less diversity compared to subset 2, as some of the former's mispredicted inputs are highly similar and somewhat redundant.
Although subset 1 has a higher misprediction rate (70\% for subset 1 vs. 40\% for subset 2), the second subset is more informative and effective for testing the DNN model. This is because it is more diverse and contains mispredicted inputs that potentially reveal more faults in the DNN model. A test set that consistently highlights the same faults in a DNN model is a waste of computational resources, particularly when faced with constraints such as limited testing budgets and high labeling costs for testing data~\cite{zohdinasab2021deephyperion,aghababaeyan2021black}. Therefore, akin to previous research on test selection in traditional software, our objective is not focused on maximizing the occurrence of failures (i.e., mispredictions) but rather to develop a test selection method that maximizes the identification of faults present in the DNN model.

Given the inherent complexity of detecting faults in DNNs, which is not as straightforward as identifying specific statements responsible for failures in traditional software, we rely on prior work for estimating faults in DNN models~\cite{aghababaeyan2021black}.
We cluster similar mispredicted inputs that exhibit common characteristics likely responsible for mispredictions. We then approximate the number of detected faults in the DNN model by counting the total number of clusters.
The proposed clustering-based fault estimation approach consists of four main steps: feature extraction, dimensionality reduction, clustering, and evaluation. The first step relies on VGG-16 to extract the feature matrix of the mispredicted inputs~\cite{aghababaeyan2021black}. 
Then, two extra features, actual and mispredicted classes related to each input, are added to the feature matrix. These two features as explained in our prior work~\cite{aghababaeyan2021black}, give us information about the misprediction behavior of the model under test and help build better clusters to reflect common causes of mispredictions. 
In the next step, dimensionality reduction techniques aim to boost the performance of clustering given the high dimensional feature space. HDBSCAN~\cite{mcinnes2017hdbscan} is then applied to cluster mispredicted inputs based on the resulting features. Final clusters are investigated with SOTA clustering evaluation metrics and through manual analysis. Empirical results have shown that (1) inputs in the same cluster are mispredicted due to the same root causes, and (2) inputs belonging to different clusters are mispredicted because of distinct faults~\cite{aghababaeyan2021black}.
Hence, although it is an approximation, this approach offers a practical and plausible method for estimating and comparing the number of detected faults across various test selection methods.

\Rev{To investigate the effectiveness of test selection approaches in a DNN and for the reasons mentioned above, we do not rely on misprediction rates as it is highly misleading. We rely instead on the fault detection rate ($FDR$) which we compute as follows: } 

\Rev{\begin{equation}\label{Eq:FDR}
 FDR(S)= \frac{\left|F_s\right| }{min (\left|S\right|, \left|F\right|) } 
\end{equation}
where $S$ is the selected test input set, $\left|F_s\right|$ is the number of faults \Rev{revealed by} $S$, $\left|S\right|$ is the test input set size, and $\left|F\right|$ is the total number of faults \Rev{revealed by the entire} dataset. The goal of our FDR metric is to measure the extent to which a test set can detect the maximum number of faults that can be found given a testing budget. This is why we compute the ratio of the number of faults actually revealed by the selected test set to the maximum potential number of faults that the subset could unveil. This potential is naturally capped by the size of the subset (i.e., testing budget) or the total number of faults in the whole dataset. We should note that testers cannot know in advance the total number of faults in the dataset, and testing budgets are determined by the resources available.
Our definition of FDR ensures that its range is always between 0 and 1, enabling us to compare the outcomes of various methods across test subsets. 
\Rev{We present two illustrative examples of how we calculate the $FDR$ using the test selection scenario depicted in Figure~\ref{fig:FaultsFig}. Given a test dataset that reveals a total number of faults $\left|F\right|$ equal to four, a test input set size $\left|S\right|$ equal to 10, and a selected subset (see \textit{Subset2}) that reveals four mispredicted inputs each belonging to a distinct fault (cluster), in this case the $FDR(S)$ is calculated as follows: $FDR(S)= \frac{4}{min(10, 4)} = 100\%$. Now, considering the same test dataset but selecting only three inputs ($\left|S\right|=3$), which include two mispredicted inputs originating from the same fault ($\left|F_s\right|=1$), the $FDR(S)$ is calculated as: $FDR(S) = \frac{1}{\min(3, 4)} = 33\%$.} The above examples illustrate that FDR represents the percentage of faults that can be detected in a subset relative to the maximum number of faults that can be detected, accounting for both the subset size and the total number of faults in the entire dataset.}

\subsection{Subject Datasets and Models}

Table~\ref{tab:Dataset} shows the combinations of datasets and models used in our experiments. For our large-scale empirical evaluation on image classification systems, we consider a set of four well-known and publicly available image recognition datasets which are MNIST~\cite{deng2012mnist}, Cifar-10~\cite{Cifar10}, Fashion-MNIST~\cite{deng2012mnist} and SVHN~\cite{netzer2011reading}.

MNIST is a dataset containing 70,000 black-and-white images of handwritten digits (60,000 for training and 10,000 for testing). Each test input is an image representing a digit from zero to nine. 
The Cifar-10 dataset contains 60,000 colored images belonging to 10 different classes (e.g., cats, dogs, airplanes). It includes 50,000 images for training and 10,000 for testing.
We also use Fashion-MNIST, containing 70,000 grayscale images of clothes (60,000 for training and 10,000 for testing). Each one of the images is associated with a label from 10 classes of fashion and clothing items. The Street View House Numbers (SVHN) is a real-world dataset that contains digit images of house numbers collected by Google Street View. It includes 99,289 images in total, including 73,257 for training and 26,032 for testing. 

We use these datasets with five state-of-the-art DNN models: LeNet1, LeNet4, LeNet5, ResNet20, and a 12-layer Convolutional Neural Network (12-layer ConvNet).
Because we compare our results with SOTA baseline approaches~\cite{pei2017deepxplore, kim2019guiding, gerasimou2020importance, gao2022adaptive, ma2021test, feng2020deepgini, Ma2018DeepGaugeMT}, we used similar combinations of models and datasets, focusing on the most widely used ones. Further, these models and datasets involve a variety of distinct inputs (in terms of classes and domain concepts) and different internal architectures, and thus constitute a good benchmark for observing key trends in DNN test selection. Finally, based on previously reported results~\cite{aghababaeyan2021black}, we provide the accuracy and the number of faults of the used models in Table~\ref{tab:Dataset}. \Rev{We should note that in our study, we intentionally trained some of our DNNs in a slightly suboptimal manner to make room for improvement in the retraining experiments conducted in RQ2. The accuracy levels achieved in our models are nonetheless consistent with existing studies on test selection for DNN models~\cite{gao2022adaptive}.}

\begin{table}[h]
    \centering
    \caption{Datasets and models used for evaluation}
    \vspace{-1em}
    \begin{tabular}{llll}
    \hline \hline
    \textbf{Dataset} & \textbf{DNN Model} & \textbf{Accuracy} & \textbf{\# Faults}  \\ \hline \hline
    
    MNIST & LeNet5 &    87.85\% & 85 \\ 
    
    &  LeNet1 &   84.5\% &  137\\ \hline
    
    \added{Fashion-MNIST} & \added{LeNet4} &     \added{88\%} & 141 \\  \hline 
    
    Cifar-10 & \raggedright 12-layer ConvNet &  82.93\% & 187 \\

    & \added{ResNet20}  &  \added{86\%} & 177 \\ \hline
     \added{SVHN} & LeNet5 &     \added{88\%} & 147 \\  \hline

    \end{tabular}
    \label{tab:Dataset}
    \vspace{-1em}
\end{table}

\subsection{Baseline Approaches}

We compare \textit{DeepGD} with two categories of baseline approaches:
(1) four white-box test selection approaches including three well-known coverage-based approaches~\cite{pei2017deepxplore}, along with Likelihood-based Surprise Adequacy (LSA) and Density-based Surprise Adequacy (DSA)~\cite{kim2019guiding}, and (2) four SOTA black-box prioritization approaches including Maximum Probability (MaxP)~\cite{ma2021test}, DeepGini~\cite{feng2020deepgini}, Adaptive Test Selection (ATS)~\cite{gao2022adaptive} and Random Selection (RS). We have selected the most prominent and recent baselines that can be applied and replicated on our models and datasets. \\

\noindent{\textbf{A- White-Box Test Selection Baselines}}

\textbf{Neuron Coverage (NC)}. It is the first coverage metric that has been proposed in the literature to test DNN models~\cite{pei2017deepxplore}. It is defined as the ratio of neurons activated by a test input to the total number of neurons in the DNN model. A neuron is activated when its activation value is greater than a predefined threshold.
    

\textbf{Neuron Boundary Coverage (NBC)}. This coverage metric measures the ratio of corner case regions that have been covered by test input(s). Corner case regions are defined as the activation values that are below or higher than the activation ranges observed during the training phase of the DNN model under test.
    
\textbf{Strong Neuron Activation Coverage (SNAC)}. Similar to the NBC metric, SNAC measures how many upper corner cases have been covered by test input(s). Upper corner cases are defined as neuron activation values that are above the activation ranges observed during the training phase of the DNN model under test.
    

\textbf{LSA and DSA}. These two metrics have been proposed by Kim \textit{et al.}~\cite{kim2019guiding} and are based on the analysis of how surprising test inputs are with respect to the training dataset. LSA uses Kernel Density Estimation (KDE)~\cite{wand1994kernel} to estimate the likelihood of seeing a test input during the training phase. According to Kim \textit{et al.}~\cite{kim2019guiding}, test inputs with higher LSA scores are preferred since they are closer to the classification boundaries. Thus, it could be considered a priority score for DNN test selection. DSA is an alternative to LSA that uses the distance between the activation traces~\cite{kim2019guiding} of new test inputs and the activation traces observed during training. \\

\noindent{\textbf{B- Black-Box Test Selection Baselines}}
      
 \textbf{DeepGini}. It is a test selection approach that prioritizes test inputs with higher uncertainty scores~\cite{feng2020deepgini}. It relies on the \textit{Gini} metric (Section~\ref{SubSec:Ginimethod}) to estimate the probability of misclassifying a test input. 
 
\textbf{MaxP}. It relies on the maximal prediction probability of the classification task to estimate the prediction confidence of the DNN model for a given input. Such method prioritizes inputs with lower confidence. The maximum probability score of a test input $x$ is defined as $MaxP(x) = 1 -\max_{i=1}^{m} P_{x_{i}}$, where $P_{x_{1 \leq i \leq m}}$ is the probability that input $x$ belongs to class $C_i$ and $m$ is the total number of classes~\cite{ma2021test}. 
Intuitively, higher MaxP scores are more likely to lead to mispredictions~\cite{ma2021test}.

\textbf{ATS}. It is a recent test selection method proposed by Gao \textit{et al.}~\cite{gao2022adaptive}. The selection is guided by a fitness function based on which test inputs are incrementally added to the final test set. The fitness function measures the difference between a test input and the currently selected test set based on computing a fault pattern coverage score. This score is obtained through analyzing the diversity of the output probability vectors of the DNN under test. They select test inputs with different fault patterns and higher uncertainty scores.  

\textbf{Random Selection} (\textbf{RS}). It is the most basic and simplest test selection method in the literature. RS consists in randomly selecting (without replacement) $\beta$ inputs from the test dataset. Each test input has the same probability of being selected. 

\subsection{Choice of Operators}

\Rev{In this work, we made modifications to the traditional NSGA-II algorithm by customizing its fundamental operators as explained in Section~\ref{Subsec:Operators}. Specifically, we introduced a novel crossover function and mutation operator. To ensure each of our customized operators had a beneficial impact on the search results, we conducted an empirical study by considering four variants of the search algorithm. Each variant focused on changing only one search operator. In the first variant of \textit{DeepGD}, we replaced the customized crossover operator with the standard version, where no reordering of the genes is applied based on their uncertainty score. In the second search variant, we replaced the customized mutation operator with the standard version, where 1\% of the genes in the selected individuals are randomly mutated, regardless of their uncertainty and diversity scores.  Finally, we introduced more mutation strategies in the two other variants. Specifically, in the third variant, we mutated the genes with the lowest uncertainty scores, aiming to explore the impact of prioritizing uncertainty in the mutation process. In the fourth variant, we focused on mutating the genes with the lowest diversity score, investigating the effect of emphasizing diversity in the mutation operation.}

\begin{table}[htbp]
    \centering
    \color{black}
    \caption{\textcolor{black}{Fault Detection Rates of\textit{ DeepGD}'s Variants}}
    \begin{tabular}{|c|c|c|c|}
        \hline
        \multirow{6}{*}{\rotatebox[]{90}{\parbox{1.9cm}{Subset size=100}}} &
        \multirow{2}{*}{\centering Method}&
        \multicolumn{2}{c|}{\ FDR} \\
        \cline{3-4}
         & & Fashion-Mnist \&  LeNet4 & Cifar-10 \& 12-layer ConvNet \\
        \hline \hline
        & Original DeepGD & \textbf{39.56\%} & \textbf{59.67\%} \\ \cline{2-4}
        & DeepGD with Simple Crossover & 31.00\% & 54.00\% \\ \cline{2-4}
        & DeepGD with Simple Mutation & 32.50\% & 49.00\% \\ \cline{2-4}
        & DeepGD with Gini-based Mutation & 38.00\% & 52.00\% \\ \cline{2-4}
        & DeepGD with GD-based Mutation & 37.00\% & 52.50\% \\
        \hline
    \end{tabular}
    \label{tab:ablation}
\end{table}

\Rev{We evaluated the different variants of \textit{DeepGD} on two models and datasets, considering a testing budget of $\beta=100$. The search results are reported in Table~\ref{tab:ablation}. Our findings demonstrate that \textit{DeepGD} consistently outperforms all search variants with its custom operators in terms of fault-revealing power across datasets and models. More specifically, reordering the genes according to their uncertainty scores when applying the crossover, helped the search converge faster and yielded better fault detection results. Moreover, incorporating both diversity and \textit{Gini} scores in the mutation process resulted in better fault detection results compared to using the default version or relying solely on \textit{Gini} or diversity scores for mutations. These results therefore justify our operators' customization choices.}

\section{Evaluation and Results}\label{Sec:Results}

We describe in this section our experimental evaluation and present in detail the obtained results. \Rev{Before addressing our research questions, it is important to note that we applied the fault estimation approach described in Section~\ref{Sec:Faults} to all models and datasets. This allowed us to identify the faults in each subject and estimate their number, which we report in Table~\ref{tab:Dataset}}.

\subsection{\textbf{RQ1. Do we find more faults than existing test selection approaches with the same testing budget?}} \label{sec:answerRQ1}

To investigate the effectiveness of test selection approaches in a DNN, existing baselines usually compare the misprediction detection rate~\cite{gao2022adaptive, feng2020deepgini}. Although the number of mispredicted inputs is a useful metric in some contexts, it is misleading for test selection in DNN-based systems since many mispredicted inputs may be redundant and caused by the same fault or root cause in the DNN model~\cite{aghababaeyan2021black}. Therefore, we compare the effectiveness of \textit{DeepGD} with the existing baselines based on the fault detection rate. However, in order to provide complete information about our analysis, the number of mispredicted inputs revealed by each approach is reported in our replication package~\cite{RepcodeDeepGD}.


\begin{table*}[ht]
    \centering
    
    \small
    \caption{The fault detection rate for each subject with test subset size $\beta=100$}
    \vspace{-1em}
    \resizebox{\textwidth}{!}{
    \begin{tabular}{|cc|    cc|   ccc|   ccc|   cc|  c c c c  |c|       }
    \hline 
          \multicolumn{2}{|c|}{}  &\multicolumn{10}{c|}{White-box}  &\multicolumn{5}{c|}{Black-box}      \\
    \cline{3-17}   
         \multirow{2}{*}{Data} &\multirow{2}{*}{Model}   
         &\multicolumn{2}{c|}{NC} &\multicolumn{3}{c|}{NBC}  &\multicolumn{3}{c|}{SNAC} 
         &\multirow{2}{*}{LSA}  &\multirow{2}{*}{DSA}   
         &\multirow{2}{*}{RS}   &\multirow{2}{*}{MaxP}  &\multirow{2}{*}{Gini}   &\multirow{2}{*}{ATS}   &\multirow{2}{*}{\textbf{DeepGD}}   \\ 
          &       &0  &0.75      &0  &0.5 &1     &0 &0.5 &1      & &        & & & & &        \\ \hline  \hline
          \multirow{2}{*}{Cifar-10} &12 ConvNet        &13\% &21\%     &13\% &12\% &12\%     &12\% &11\% &13\%    &31\% &35\%  
           &14\% &55\% &54\% &53\%       &\textbf{59\%}    \\ 
                                   &ResNet20      &10\%  &10\%      &11\% &17\% &16\%     &11\% &17\% &16\%    &42\% &37\%
           &11\% &52\% & 56\% &50\%       &\textbf{57\%}  \\  \cline{1-17}
          \multirow{2}{*}{MNIST}   &LeNet1        &25\% &15\%     &16\% &19\% &12\%     &16\% &18\% &12\%    &31\% &23\%
           &12\% & 41\% &28\% &40\%       &\textbf{42\%}   \\ 
                                   &LeNet5        &16\% &13\%     &13\% &17\% &16\%     &14\% &19\% &16\%    &29\% &33\%
           &13\% &35\% &34\% &36\%       &\textbf{40\%}    \\  \cline{1-17}
                          Fashion  &LeNet4         &5\% &18\%     &14\% &15\% &14\%     &14\% &15\% &14\%    &17\% &34\%
           &10\% &\textbf{39\%} &\textbf{39\%} &32\%       &\textbf{39\%}  \\  \cline{1-17}
                          SVHN     &LeNet5        &9\% &4\%     &11\% &12\% &11\%     &11\% &12\% &11\%  &13\% &19\%
           &11\% &43\% &43\% &44\%       &\textbf{47\%}   \\  \cline{1-17}    
         
    \end{tabular}
    }
    \label{tab:FaultRateResults100}
\end{table*}

\begin{table*}[ht]
    \centering
    \caption{The fault detection rate for each subject with test subset size $\beta=300$}
    \small
    \vspace{-1em}
    \resizebox{\textwidth}{!}{
    \begin{tabular}{|cc|    cc|   ccc|   ccc|   cc|  c c c c  |c|   }
    \hline 
          \multicolumn{2}{|c|}{}  &\multicolumn{10}{c|}{White-box}  &\multicolumn{5}{c|}{Black-box}      \\
    \cline{3-17}   
         \multirow{2}{*}{Data} &\multirow{2}{*}{Model}   
         &\multicolumn{2}{c|}{NC} &\multicolumn{3}{c|}{NBC}  &\multicolumn{3}{c|}{SNAC} 
         &\multirow{2}{*}{LSA}  &\multirow{2}{*}{DSA}   
         &\multirow{2}{*}{RS}   &\multirow{2}{*}{MaxP}  &\multirow{2}{*}{Gini}   &\multirow{2}{*}{ATS}   &\multirow{2}{*}{\textbf{DeepGD}}   \\ 
          &       &0  &0.75      &0  &0.5 &1     &0 &0.5 &1      & &        & & & & &        \\ \hline  \hline
        \multirow{2}{*}{Cifar-10} &12 ConvNet   &19\% &28\%     &21\% &22\% &23\%     &22\% &22\% &22\%     &35\% &37\%
           &22\% &54\% &53\% &50\%       &\textbf{57\%} \\   
                     &ResNet20           &15\% &17\%     &17\% &17\% &19\%     &17\% &17\% &19\%    &49\% &45\%
           &19\% &56\% &58\% &52\%       &\textbf{59\%} \\  \cline{1-17}  
        \multirow{2}{*}{MNIST}   &LeNet1  &25\% &25\%     &33\% &28\% &27\%     &36\% &29\% &27\%    &40\% &40\%
           &25\% &53\% & 44\% & 53\%     &\textbf{54\%} \\    
                     &LeNet5             &34\% &33\%     &32\% &33\% &30\%     &32\% &30\% &30\%    &54\% &52\%
           &24\% &63\% &59\% &62\%       &\textbf{64\%} \\  \cline{1-17}   
           Fashion  &LeNet4              &8\% &25\%     & 19\% & 23\% &23\%     &19\% &23\% &23\%    &35\% &45\%
           &17\% &51\% &47\% &49\%       &\textbf{53\%} \\  \cline{1-17}   
          SVHN     &LeNet5               &11\% &13\%     &17\% &16\% &18\%     &17\% &16\% &18\%   &19\% &36\%
           &17\% &58\% &61\% &58\%       &\textbf{62\%} \\  \cline{1-17}
          
         
    \end{tabular}
    }
    \label{tab:FaultRateResults300}
\end{table*}

\begin{table*}[ht]
    \centering
    \color{black}
    \small
    \caption{The fault detection rate for each subject with test subset size $\beta=500$}
    \vspace{-1em}
    \resizebox{\textwidth}{!}{
    \begin{tabular}{|cc|    cc|   ccc|   ccc|   cc|  c c c c  |c|       }
    \hline 
    
          \multicolumn{2}{|c|}{}  &\multicolumn{10}{c|}{White-box}  &\multicolumn{5}{c|}{Black-box}      \\
    \cline{3-17}   
         \multirow{2}{*}{Data} &\multirow{2}{*}{Model}   
         &\multicolumn{2}{c|}{NC} &\multicolumn{3}{c|}{NBC}  &\multicolumn{3}{c|}{SNAC} 
         &\multirow{2}{*}{LSA}  &\multirow{2}{*}{DSA}   
         &\multirow{2}{*}{RS}   &\multirow{2}{*}{MaxP}  &\multirow{2}{*}{Gini}   &\multirow{2}{*}{ATS}   &\multirow{2}{*}{\textbf{DeepGD}}   \\ 
          &       &0  &0.75      &0  &0.5 &1     &0 &0.5 &1      & &        & & & & &        \\ \hline  \hline
          \multirow{2}{*}{Cifar-10} &12 ConvNet        &30\% &	34\% &	32\% &	33\% &	30\% &	30\% &	31\% &	29\% &	45\% &	51\% &	33\% &	67\% &	65\% &	64\% &	\textbf{70\%}   \\ 
                                   &ResNet20      &28\% &	28\% &	28\% &	29\% &	30\% &	28\% &	29\% &	30\% &	61\% &	63\% &	27\% &	68\% &	71\% &	63\% &	\textbf{73\%}  \\  \cline{1-17}
          \multirow{2}{*}{MNIST}   &LeNet1        &29\% &	31\% &	43\% &	39\% &	36\% &	42\% &	38\% &	36\% &	52\% &	58\% &	34\% &	66\% &	58\% &	65\% &	\textbf{67\%}   \\ 
                                   &LeNet5        &39\% &	42\% &	41\% &	46\% &	40\% &	42\% &	42\% &	40\% &	65\% &	65\% &	38\% &	73\% &	72\% &	71\% &	\textbf{74\%}    \\  \cline{1-17}
                          Fashion  &LeNet4         &13\% &	28\% &	28\% &	29\% &	27\% &	28\% &	29\% &	27\% &	46\% &	54\% &	27\% &	\textbf{70\%} &	65\% &	64\% &	\textbf{70\%}  \\  \cline{1-17}
                          SVHN     &LeNet5        &14\% &	22\% &	24\% &	22\% &	25\% &	24\% &	22\% &	25\% &	27\% &	46\% &	26\% &	77\% &	76\% &	72\% &	\textbf{78\%}   \\  \cline{1-17}    
         
    \end{tabular}
    }
    \label{tab:FaultRateResults500}
\end{table*}

Detailed results of the fault detection rate for six subjects (resulting from the combination of datasets and DNN models), and for different sizes of the test input set $\beta \in \{100, 300, 500\}$, are shown in Table~\ref{tab:FaultRateResults100}, Table~\ref{tab:FaultRateResults300}, and Table~\ref{tab:FaultRateResults500}, respectively. To provide a more comprehensive evaluation of our results, we also report the average number of faults covered by each selection method later in Tables \ref{tab: tabapp1} through \ref{tab: tabapp3} of Appendix 1. The results follow similar trends and the final conclusion is consistent for all test subset sizes. Because of randomness in \textit{DeepGD}, ATS, and random selection, we re-executed each of them 10 times and reported the corresponding average fault detection rates and average number of detected faults (see Appendix 1).
Our results show that all of the test selection approaches guided by coverage metrics are ineffective in detecting faults. Even after repeating the experiment with various parameters as suggested in their original papers, the results are similar. The fault detection rate of NC, for example, was between 4\% and 25\% for subset sizes of 100. SNAC and NBC also showed poor fault detection rates, which vary between 11\% and 19\% for the same subset size. Compared with LSA and DSA, both \textit{DeepGD} and other black-box prioritization selection methods showed a much higher fault detection rate for all subjects. \Rev{Considering our previous experiments~\cite{aghababaeyan2021black}, where we investigated white-box coverage metrics and reported that they are not correlated to faults, it is not surprising that these approaches are unlikely to perform as effectively as black-box approaches in the selection of test inputs with high fault-revealing power.}
Overall, it can be concluded that black-box approaches are more effective in detecting faults than white-box approaches. 
\textit{DeepGini} and MaxP fare the same as \textit{DeepGD} for two different subjects with a test subset size of 100. However, for other subjects, \textit{DeepGD} achieves a higher fault detection rate than these two approaches. For larger test subset sizes ($\beta \in \{300, 500\}$), \textit{DeepGD} performs better than other black-box baselines in all subjects, with a fault detection rate between 53\% and 78\%.

We should note that the second-best approach is not consistently the same across subjects and sizes, therefore strengthening our conclusion that \textit{DeepGD} is the only solution we can confidently recommend regardless of the model, dataset, and test set size. For example, with $\beta=100$, compared to black-box SOTA baselines (ATS, MaxP, Gini), \textit{DeepGD} is, on average, 2 pp better (with a maximum of 4 pp across all models and datasets) than the second-best black-box alternative and 7 pp better (with a maximum of 14 pp across all models and datasets) than the fourth-best black-box alternative (excluding random selection which consistently showed poor performance). Since we cannot predict beforehand how a given alternative will fare compared to the others for a dataset and model, this implies that the effect of selecting another technique than \textit{DeepGD} is potentially significant as we may end up using the worst alternative. For example, if we rely on the inputs selected by \textit{DeepGini} to test LeNet1 on the MNIST dataset ($\beta=300$), we will only reach a fault detection rate of 44\% instead of 54\% with \textit{DeepGD}. Further, we also report that \textit{DeepGD} is on average 15 pp better than the best white-box test selection approach and 41 pp better than the worst white-box alternative across all models and datasets.

We performed the statistical analysis using the Wilcoxon signed-rank test~\cite{macfarland2016wilcoxon}, with a significance level of $\alpha = 0.05$, to investigate whether \textit{DeepGD} significantly outperforms each SOTA baseline in terms of fault-revealing power. The Wilcoxon signed-rank test, a non-parametric test, is used to compare the medians of continuous variables for paired samples and is employed when the data does not follow a normal distribution, as in our case. We employ a paired test as we compare the fault-revealing power of methods for a given subset size and model. To conduct the statistical tests, we collected data about the performance of each method (DeepGD, ATS, GINI, and MaxP) across the 18 different combinations of datasets, models, and subset sizes. For each SOTA baseline, we thus compared pairs of fault detection rates reported for DeepDG and the selected baseline across 18 different combinations. Results showed that all p-values are lower than 0.05, indicating that \textit{DeepGD} significantly surpasses all SOTA baselines in finding more faults in DNNs.

\begin{tcolorbox} \textbf{Answer to RQ1:} 
\textit{DeepGD} outperforms both white-box and black-box test selection approaches for DNNs, in terms of detecting distinct faults given the same testing budget. Further, the second-best black-box approach after \textit{DeepGD} is not consistently the same. 
\end{tcolorbox}

\subsection{\textbf{RQ2. Do we more effectively guide the retraining of DNN models with our selected inputs than with baselines?}} 

Having examined the effectiveness of \textit{DeepGD} and other baselines in selecting test input sets with high fault-revealing power, we next focus on the extent to which the test selection approaches can help select data to effectively retrain the DNN models under test. 
We only consider black-box test selection approaches in this research question, as they showed much better performance than white-box test selection baselines. We consider the same models and datasets as in the previous experiment. We augment the original training dataset with the test input selected in RQ1 with the size of 300 by \textit{DeepGD}, ATS, DeepGini, and MaxP, respectively, to retrain the DNN model. We measure the accuracy of the retrained model on both the whole test dataset and a newly generated dataset that is obtained by applying five realistic image transformations to the original test dataset. The latter allows for a fairer and more complete comparison between our proposed method and the other baseline approaches since none of the inputs in the generated dataset were used for retraining the model.
We leveraged the datasets generated by Gao \textit{et al.}~\cite{gao2022adaptive}, a recent work in test selection, that we consider as one of our baselines. The authors applied various transformations to the MNIST, Fashion-MNIST, and Cifar-10 test datasets to generate more valid test inputs. However, since the SVHN dataset was not included in their experiments, to ensure consistency, we generated new test inputs for this dataset as well by applying their transformations with identical settings, ensuring the validity of generated inputs. These transformations include sheering, rotating, zooming, and changing the brightness and the blurriness of all images in each dataset. We should note that the size of the newly generated test dataset is five times larger than the original test dataset since we apply five image transformations on each original test input. For each black-box test selection approach, we measure the accuracy improvement of the retrained models on both the original test dataset and the generated one. We report the results in Table~\ref{tab:AccuracyRetraining}. 
We acknowledge that studying the accuracy improvement of the retrained models on the generated datasets is more suitable in our context since it does not contain any of the inputs used for retraining. But we choose to also report improvements with the original test dataset to verify that the retraining process did improve model accuracy. Note that we only select a small part of the original test dataset to retrain the model (only 300 out of 10,000 to 26,000 inputs). Because of the randomness in \textit{DeepGD} and ATS, we re-executed each of them five times on the different subjects. For each execution, we retrained each subject with the selected test input set and reported the corresponding average model accuracy improvements.

\begin{table*}[ht]
\centering
\caption{DNNs accuracy improvements after retraining with the selected test input sets.}

\label{tab:AccuracyRetraining}
\begin{tabular}{cccccclllc}
\cline{3-10}
 & \multicolumn{1}{c|}{} & \multicolumn{4}{c||}{Accuracy imp. on orig. test dataset} & \multicolumn{4}{c|}{Accuracy imp. on generated test dataset}  \\ \cline{1-10}
\multicolumn{1}{|c|}{Data} & \multicolumn{1}{c|}{Model} & MaxP & Gini & \multicolumn{1}{c|}{ATS} & \multicolumn{1}{c||}{\textbf{DeepGD}} & \multicolumn{1}{c}{MaxP} & \multicolumn{1}{c}{Gini} & \multicolumn{1}{c|}{ATS} & \multicolumn{1}{c|}{\textbf{DeepGD}}   \\ \hline \hline
\multicolumn{1}{|c|}{\multirow{2}{*}{Cifar-10}} & \multicolumn{1}{c|}{12 ConvNet} & 3.52\% & 2.09\% & \multicolumn{1}{c|}{2.12\%} & \multicolumn{1}{c||}{\textbf{3.53\%}} & 5.35\% & 5.12\% & \multicolumn{1}{l|}{4.62\%} & \multicolumn{1}{l|}{\textbf{6.85\%}}   \\ \cline{2-10}
\multicolumn{1}{|c|}{} & \multicolumn{1}{c|}{ResNet20} & 1.66\% & 0.74\% & \multicolumn{1}{c|}{2.03\%} & \multicolumn{1}{c||}{\textbf{2.50\%}} & 4.12\% & 3.45\% & \multicolumn{1}{l|}{5.97\%} & \multicolumn{1}{l|}{\textbf{6.18\%}}   \\ \cline{1-10}
\multicolumn{1}{|c|}{\multirow{2}{*}{MNIST}} & \multicolumn{1}{c|}{LeNet1} & 10.39\% & 10.40\% & \multicolumn{1}{c|}{10.40\%} & \multicolumn{1}{c||}{\textbf{11.10\%}} & 13.18\% & 13.24\% & \multicolumn{1}{l|}{13.19\%} & \multicolumn{1}{l|}{\textbf{14.76\%}}   \\ \cline{2-10}
\multicolumn{1}{|c|}{} & \multicolumn{1}{c|}{LeNet5} & 7.94\% & 7.91\% & \multicolumn{1}{c|}{7.92\%} & \multicolumn{1}{c||}{\textbf{9.26\%}} & 13.02\% & 12.92\% & \multicolumn{1}{l|}{13.08\%} & \multicolumn{1}{l|}{\textbf{15.82\%}}   \\ \cline{1-10}
\multicolumn{1}{|c|}{Fashion} & \multicolumn{1}{c|}{LeNet4} & 3.94\% & 3.94\% & \multicolumn{1}{c|}{3.91\%} & \multicolumn{1}{c||}{\textbf{4.17\%}} & 7.33\% & 7.41\% & \multicolumn{1}{l|}{7.62\%} & \multicolumn{1}{l|}{\textbf{7.69\%}}   \\ \cline{1-10}
\multicolumn{1}{|c|}{SVHN} & \multicolumn{1}{c|}{LeNet5} & 0.96\% & 0.99\% & \multicolumn{1}{c|}{0.93\%} & \multicolumn{1}{c||}{\textbf{1.24\%}} & 0.61\% & 0.54\% & \multicolumn{1}{l|}{0.63\%} & \multicolumn{1}{l|}{\textbf{1.26\%}}   \\ \cline{1-10}
 &  &  &  &  &  & \multicolumn{1}{c}{} & \multicolumn{1}{c}{} & \multicolumn{1}{c}{} & \multicolumn{1}{c}{} 
\end{tabular}
\end{table*}

\begin{table*}[ht]
\centering
\footnotesize
\caption{Optimization effectiveness compared to retraining with all candidate tests}
\label{tab:OptimizationEffectiveness}
\begin{tabular}{ccc|cccc||c|cccc|}
\cline{4-7} \cline{9-12}
 &  &  & \multicolumn{4}{c|}{Opt effectiveness on orig. test dataset (T)} &  & \multicolumn{4}{c|}{Opt effectiveness on generated dataset (G)} \\ \hline
\multicolumn{1}{|c|}{Data} & \multicolumn{1}{c|}{Model} &\begin{tabular}[c]{@{}l@{}}Max imp.\\ (100\% T)\end{tabular} & MaxP & Gini & \multicolumn{1}{c|}{ATS} & \textbf{DeepGD} & \begin{tabular}[c]{@{}l@{}}Max imp.\\ (100\% G)\end{tabular} & MaxP & Gini & \multicolumn{1}{c|}{ATS} & \textbf{DeepGD} \\ \hline \hline
\multicolumn{1}{|c|}{\multirow{2}{*}{Cifar-10}} & \multicolumn{1}{c|}{12 ConvNet} & 16.36\% & 21.51\% & 12.78\% & \multicolumn{1}{c|}{12.96\%} & \textbf{21.58\%} & 28.98\% & 18.47\% & 17.65\% & \multicolumn{1}{c|}{15.94\%} & \textbf{23.65\%} \\ \cline{2-12} 
\multicolumn{1}{|c|}{} & \multicolumn{1}{c|}{ResNet20} & 9.99\% & 16.62\% & 7.41\% & \multicolumn{1}{c|}{20.32\%} & \textbf{25.03\%} & 23.33\% & 17.65\% & 14.80\% & \multicolumn{1}{c|}{25.59\%} & \textbf{26.47\%} \\ \hline
\multicolumn{1}{|c|}{\multirow{2}{*}{MNIST}} & \multicolumn{1}{c|}{LeNet1} & 11.10\% & 93.60\% & 93.72\% & \multicolumn{1}{c|}{93.69\%} & \textbf{100\%} & 23.80\% & 55.39\% & 55.62\% & \multicolumn{1}{c|}{55.43\%} & \textbf{62.04\%} \\ \cline{2-12} 
\multicolumn{1}{|c|}{} & \multicolumn{1}{c|}{LeNet5} & 9.52\% & 83.40\% & 83.08\% & \multicolumn{1}{c|}{83.19\%} & \textbf{97.27\%}& 22.78\% & 57.15\% & 56.72\% & \multicolumn{1}{c|}{57.41\%} & \textbf{69.46\%} \\ \hline
\multicolumn{1}{|c|}{Fashion} & \multicolumn{1}{c|}{LeNet4} & 12.00\% & 32.83\% & 32.83\% & \multicolumn{1}{c|}{32.58\%} & \textbf{34.75\%} & 23.72\% & 30.91\% & 31.23\% & \multicolumn{1}{c|}{32.10\%} & \textbf{32.41\%} \\ \hline
\multicolumn{1}{|c|}{SVHN} & \multicolumn{1}{c|}{LeNet5} & 1.25\% & 77.01\% & 79.79\% & \multicolumn{1}{c|}{74.84\%} & \textbf{99.90\%} & 8.92\% & 6.87\% & 6.06\% & \multicolumn{1}{c|}{7.04\%} & \textbf{14.15\%} \\ \hline \hline
\multicolumn{2}{|c|}{Average} & 10.04\% & 54.16\% & 51.60\% & \multicolumn{1}{c|}{52.93\%} & \textbf{63.09\%} & 21.92\% & 31.07\% & 30.35\% & \multicolumn{1}{c|}{32.25\%} & \textbf{38.03\%} \\ \hline
\end{tabular}
\end{table*}

As described in Table~\ref{tab:AccuracyRetraining}, we found that \textit{DeepGD} provides better guidance for retraining DNN models than the other black-box test selection approaches. It consistently outperforms ATS, DeepGini, and MaxP in terms of accuracy improvements across all models and datasets. Similar to the previously reported results in RQ1, we found that the second-best approach for guiding retraining is not consistently the same across all subjects. For example, ATS was the second-best approach for retraining ResNet20 with Cifar-10, while it was the third-best approach for retraining LeNet5 with SVHN. 

To further investigate the effectiveness of \textit{DeepGD} and the other black-box baselines in retraining DNN models, we also computed their \textit{optimization effectiveness} by accounting for the maximum possible accuracy improvement. We therefore retrain the model with the entire original test dataset and report the best accuracy that can be achieved by retraining. Then, for each test selection method, we calculate its optimization effectiveness, defined as the ratio of (1) the accuracy improvement when retaining the model with only 300 selected inputs from the original test dataset (Table~\ref{tab:AccuracyRetraining}) to (2) the accuracy improvement when retaining the model with the entire original test dataset. 
We also repeat the above experiment for the generated test data set to once again get a fairer comparison of the test selection approaches and to obtain better generalizability for our results. More specifically, we retrain the original model by adding all generated inputs to the training dataset to achieve the highest accuracy possible. Then, we calculate the corresponding optimization effectiveness which is the ratio of (1) the accuracy improvement when retaining the model with only 300 inputs from the original test dataset (Table~\ref{tab:AccuracyRetraining}) to (2) the accuracy improvement when retaining the model with the entire generated test dataset. Because of the randomness in \textit{DeepGD} and ATS, we re-executed each of them five times on the different subjects and reported the corresponding average results in Table~\ref{tab:OptimizationEffectiveness}. 

We found that \textit{DeepGD} consistently outperforms other black-box test selection approaches in terms of optimization effectiveness. Compared to black-box SOTA baselines, \textit{DeepGD} is, on average, 8.93 pp better (with a maximum of 20.11 pp) than the second-best black-box alternative and 11.49 pp better (with a maximum of 25.06 pp) than the worst black-box alternative. As for RQ1, it is worth noting that since the performance of alternatives is not consistent across datasets and models, selecting one of them may yield the worst results. 
We also report the average optimization effectiveness on the generated test datasets. Gini, MaxP and ATS yield 30.35\%, 31.07\%, and 32.25\% respectively, while \textit{DeepGD} achieves 38.03\%. 
In other words, with only 300 test inputs, from the original test dataset and selected by \textit{DeepGD} for retraining, we were able to reach 38.03\% of the maximum achievable accuracy by retraining with all generated test dataset. 
Similar to RQ1, we performed a statistical analysis using Wilcoxon signed-rank tests, with a significance level of 0.05, to investigate whether \textit{DeepGD} significantly outperforms the selected black-box test selection approaches in terms of optimization effectiveness across all subjects. We found all p-values to be less than 0.05, indicating that \textit{DeepGD} is significantly better than the selected baselines in retraining DNN models.  
In other words, results show that \textit{DeepGD} can select a small, more informative subset from a large unlabeled dataset to effectively retrain DNN models and minimize the labeling costs. 
Selecting diverse inputs with high uncertainty scores not only helps at detecting more faults in the DNN model, but also significantly improves the accuracy of the model through retraining. 

\begin{tcolorbox}
\textbf{Answer to RQ2:} \textit{DeepGD} provides better guidance than black-box alternatives for retraining DNN models. It consistently and statistically outperforms other black-box test selection approaches in terms of accuracy improvement, in absolute terms and relatively to the maximum achievable improvement.   
\end{tcolorbox}

\subsection{\textbf{RQ3. Can DeepGD select more diverse test input sets?}}

\begin{figure*}[ht]
\begin{minipage}{.33\textwidth}
\centering
\includegraphics[scale=0.33]{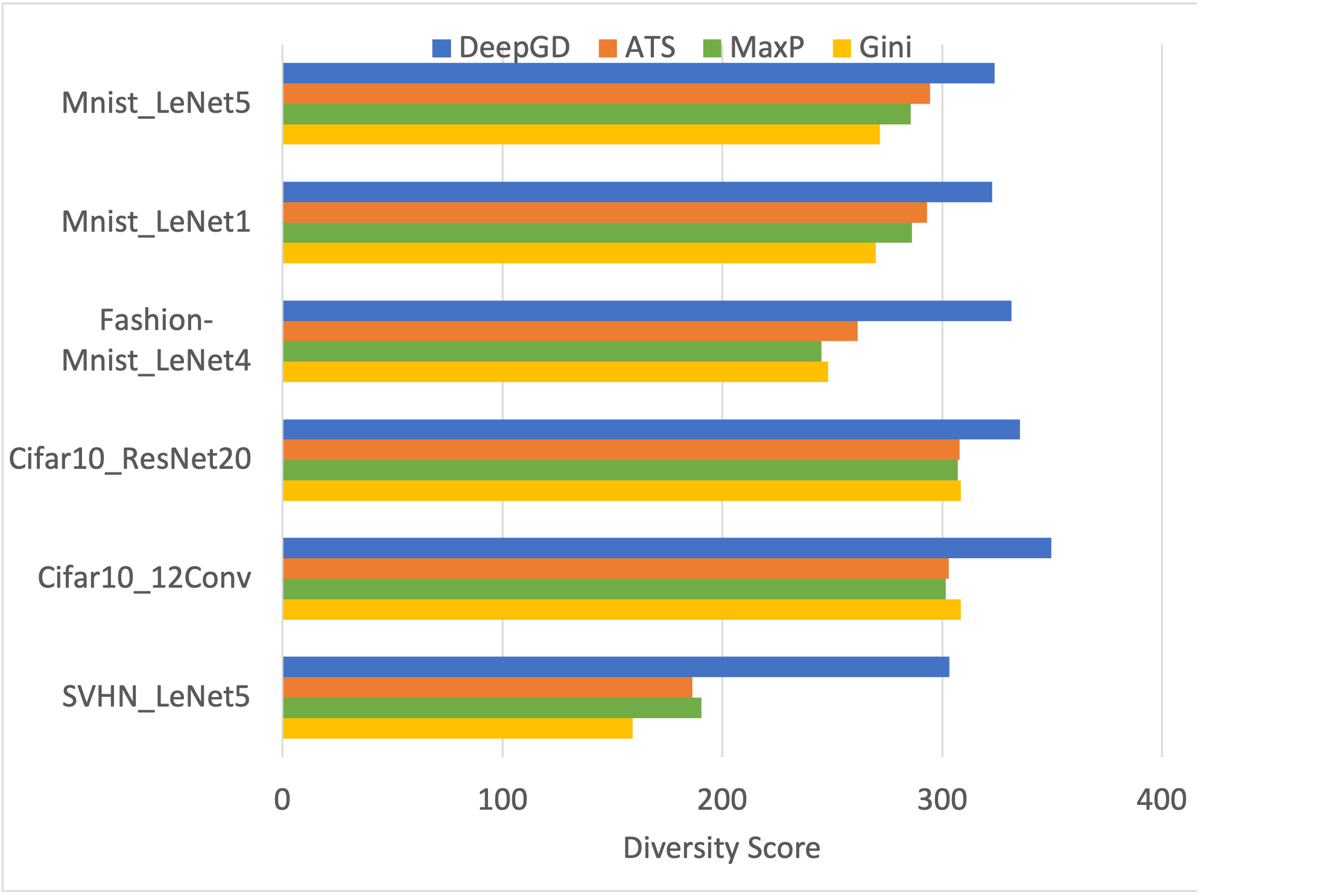}\\
\caption{\textcolor{black}{GD scores for subsets size=100}}
\label{Fig:GD100}
\end{minipage}
\begin{minipage}{.33\textwidth}
\centering
\includegraphics[scale=0.33]{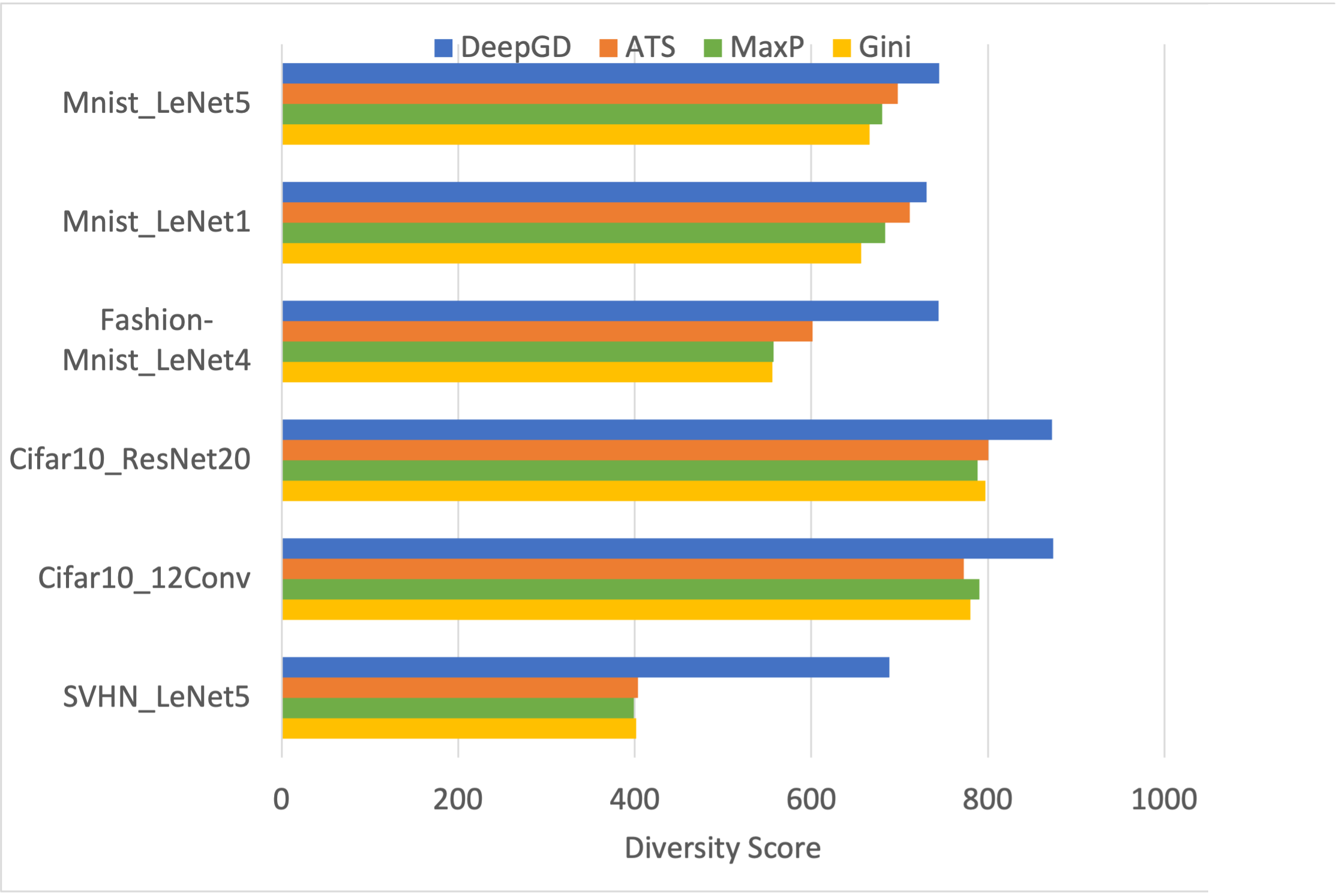}\\
\caption{\textcolor{black}{GD scores for subsets size=300}}
\label{Fig:GD300}
\end{minipage}
\begin{minipage}{.3\textwidth}
\centering
\includegraphics[scale=0.33]{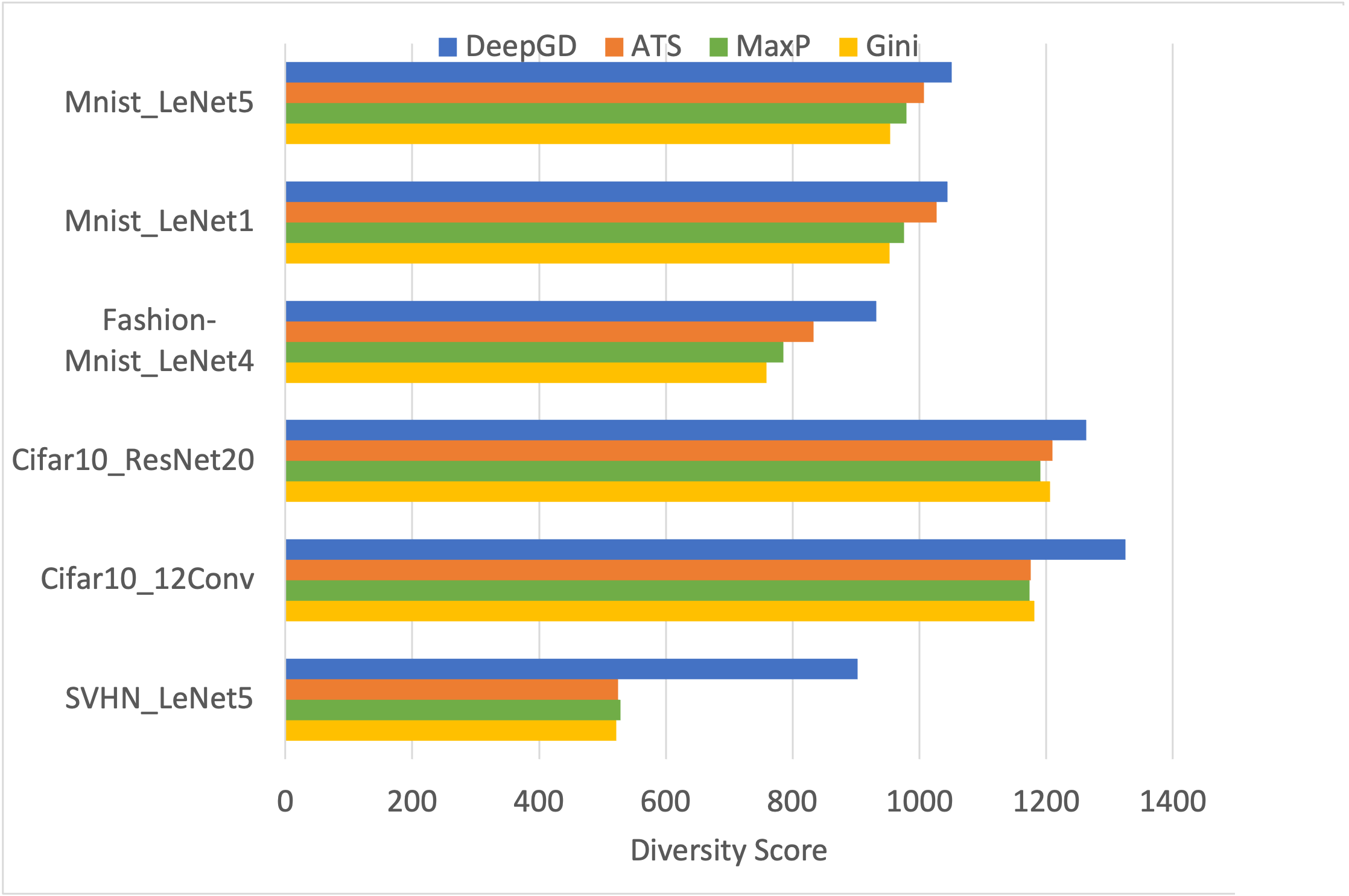}\\
\caption{\textcolor{black}{GD scores for subsets size=500}}
\label{Fig:GD500}
\end{minipage}
\end{figure*}

\Rev{To answer this research question, we measured the diversity score of the selected test input sets for \textit{DeepDG} and the other black-box baselines. We relied on the geometric diversity metric since we demonstrated in our prior work~\cite{aghababaeyan2021black} its construct validity in measuring the actual diversity of input sets. 
We used the same subjects, subset sizes and test input sets that were selected in RQ1. Given the inherent randomness in \textit{DeepGD} and ATS, we reported the average diversity scores based on 10 executions. The results, presented in Figures~\ref{Fig:GD100},~\ref{Fig:GD300} and~\ref{Fig:GD500}, consistently demonstrate that \textit{DeepGD} outperforms the baseline approaches in selecting diverse input sets. Across all subjects and subset sizes, \textit{DeepGD} consistently obtained the highest diversity scores. Additionally, ATS emerged as the second-best approach for selecting diverse input sets in 12 out of 18 configurations, while \textit{Gini} performed in general the poorest, ranking last in 11 out of 18 configurations.}

\Rev{We also performed a statistical test analysis using the Wilcoxon signed-rank test, with a significance level of 0.05, to investigate whether \textit{DeepGD} significantly outperforms each baseline approach in selecting diverse test input sets.
Similarly to RQ1 and RQ2, we performed a paired test across the different combinations of datasets, models, and subset sizes. For each baseline approach, we compared pairs of diversity scores reported for \textit{DeepGD} and the selected baseline across the 18 different configurations. Results show that all p-values are below 0.05, indicating that \textit{DeepGD} significantly outperforms all SOTA baselines in selecting more diverse test input sets.}

\begin{tcolorbox}
\Rev{\textbf{Answer to RQ3:} \textit{DeepGD} provides better guidance than black-box alternatives for selecting diverse test input sets. It consistently and statistically outperforms other black-box test selection baselines in terms of diversity. }
\end{tcolorbox}

\subsection{\textbf{RQ4. How do DeepGD and baseline approaches compare in terms of computation time?}}

\Rev{To address this research question, we compared the execution time of \textit{DeepGD} with baseline approaches and analyzed its evolution with different subset sizes. Similar to RQ2, we focused on black-box test selection approaches (excluding random selection) as they demonstrated better performance compared to white-box test selection baselines. The subjects and subset sizes used were consistent with our previous experiments. For each subject and subset size, we executed \textit{DeepGD} and the selected baseline approaches 10 times and measured the computation times. This repetition allowed us to account for random variations in execution time. }

\begin{figure}[h]
\centering
\includegraphics[width=\textwidth]{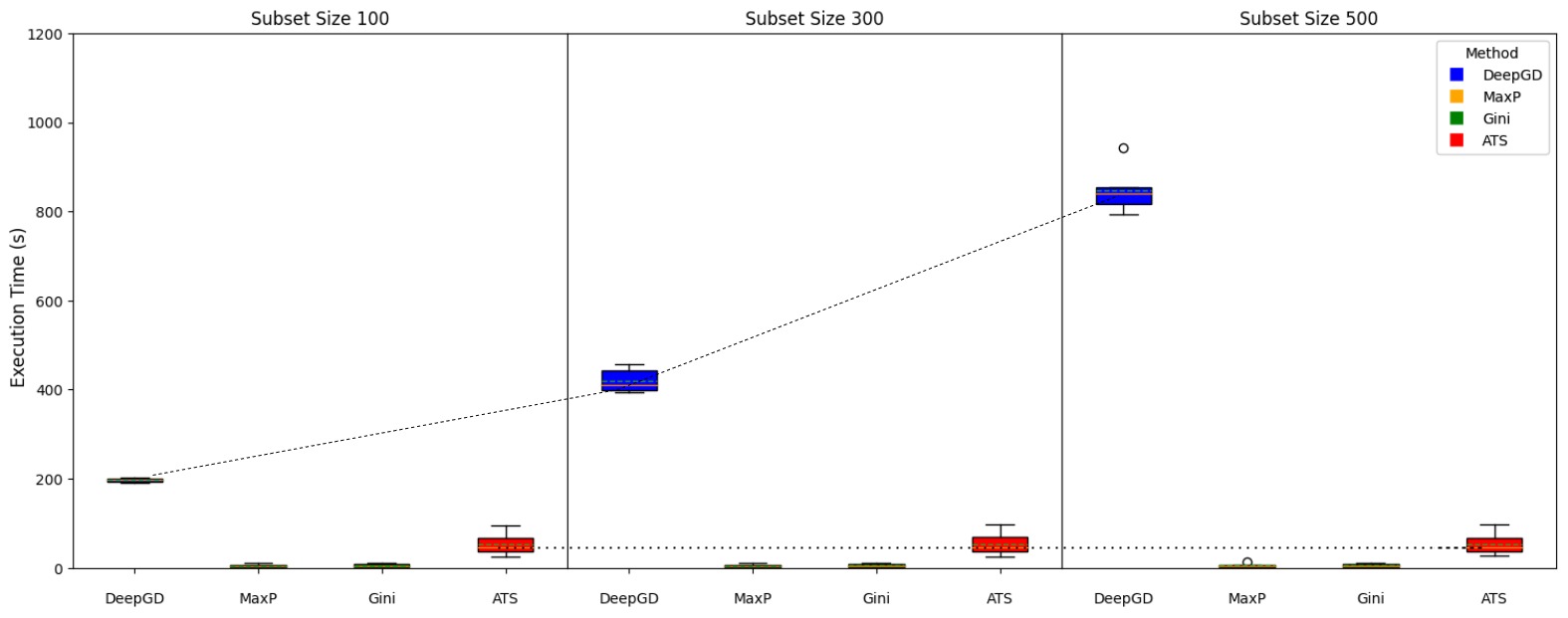}
\caption{Evolution of the execution time over the subset sizes}
\label{fig:time}
\end{figure}

\Rev{The results, presented in Figure~\ref{fig:time}, illustrate the change in computation times as the size of the input sets increased. Our findings indicate that all selected approaches were computationally efficient. \textit{Gini} and MaxP exhibited similar execution times, approximately five seconds, and demonstrated the best performance in terms of computation time. In contrast, \textit{DeepGD}, since it relies on a search algorithm, had higher execution times ranging from 200s to 850s across all subjects and subset sizes. Furthermore, we observed that, unlike the other baseline approaches, the execution time of \textit{DeepGD} increases with subset size. However, it is important to note that such execution times remain practically acceptable since, in DNN testing, (1) the labeling cost of test inputs is far more expensive than the computation cost of the search, and (2) test selection is neither frequent nor a real-time task. Despite the longer execution times, the better fault detection performance and retraining guidance provided by \textit{DeepGD} therefore outweigh the increase in computation time in comparison to the baselines.}

\begin{tcolorbox}
\Rev{\textbf{Answer to RQ4:} 
None of the black-box test selection approaches are computationally expensive. Though \textit{DeepGD} yields longer execution times (a matter of minutes), these remain practical, even for large test subsets, given that test selection is not a frequent task performed under tight time constraints. }
\end{tcolorbox}

\subsection{\textbf{Discussions}}\label{Sec:Discussions}

Based on our experimental results, we show that \textit{DeepGD} provides better guidance than existing baselines for selecting test inputs with high-fault revealing power. The second-best approaches for test selection and model retraining are not consistently the same across all models and datasets. This reinforces our conclusion that \textit{DeepGD} is the only solution that we can confidently recommend regardless of the model, dataset, and test set size. Indeed, selecting diverse test inputs with high uncertainty scores not only enables higher fault detection, but also provides more effective guidance for retraining DNN models. Although results are encouraging, and though this remains to be further investigated, we conjecture that \textit{DeepGD} is particularly useful when datasets are more redundant, a situation that is often observed in real-world DNN testing scenarios, especially with massive generated datasets. Selecting diverse test input sets with \textit{DeepGD} is then expected to help reduce redundancy by selecting more informative inputs to be labeled in order to test and retrain DNN models. 


Since \textit{DeepGD} relies on genetic search, it entails some degree of randomness in the final selected test sets. We therefore studied the stability performance of \textit{DeepGD} by running the tool 10 times with each combination of dataset, model, and subset size that we have previously used in our experiments. We report the minimum, maximum, and standard deviation of FDR results in Table~\ref{Tab:Stability}. As shown in the table, the standard deviations of the FDRs reported by \textit{DeepGD} are consistently low, ranging from 0.008 to 0.021. This indicates that \textit{DeepGD} consistently delivers stable and thus reliable results across subjects and subset sizes over multiple runs. 
\Rev{We also found that the faults revealed by \textit{DeepGD} are largely consistent across multiple runs. This can be attributed to the use of the \textit{Gini} score as one of the fitness functions to guide the search towards test inputs with high uncertainty scores.  }

\begin{table}[ht]
\centering
\small
\caption{Stability of \textit{DeepDG's} fault detection rates.}
\begin{tabular}
{cc|cccc||cccc|ccc}
\cline{3-6}\cline{7-10}
                                                &            & \multicolumn{4}{c||}{Size =100}                                                          & \multicolumn{4}{c|}{Size =300}                                                          &  &  &  \\ \cline{1-10}
\multicolumn{1}{|c|}{Data}                      & Model      & \multicolumn{1}{c|}{Ave. FDR} & \multicolumn{1}{c|}{Min. FDR}  & \multicolumn{1}{c|}{Max. FDR}  & Std.  & \multicolumn{1}{c|}{Ave.} & \multicolumn{1}{c|}{Min. FDR}  & \multicolumn{1}{c|}{Max. FDR}  & Std. &  &  &  \\ \cline{1-10}
\multicolumn{1}{|c|}{\multirow{2}{*}{Cifar-10}} & 12 ConvNet & \multicolumn{1}{c|}{59\%} & \multicolumn{1}{c|}{57\%} & \multicolumn{1}{c|}{62\%} & 0.019 & \multicolumn{1}{c|}{57\%} & \multicolumn{1}{c|}{55\%} & \multicolumn{1}{c|}{58\%} & 0.009 &  &  &  \\ \cline{2-10}
\multicolumn{1}{|c|}{}                          & ResNet20   & \multicolumn{1}{c|}{57\%} & \multicolumn{1}{c|}{55\%} & \multicolumn{1}{c|}{60\%} & 0.013 & \multicolumn{1}{c|}{59\%} & \multicolumn{1}{c|}{57\%} & \multicolumn{1}{c|}{60\%} & 0.009 &  &  &  \\ \cline{1-10}
\multicolumn{1}{|c|}{\multirow{2}{*}{MNIST}}    & LeNet1     & \multicolumn{1}{c|}{42\%} & \multicolumn{1}{c|}{40\%} & \multicolumn{1}{c|}{45\%} & 0.017 & \multicolumn{1}{c|}{54\%} & \multicolumn{1}{c|}{51\%} & \multicolumn{1}{c|}{54\%} & 0.009 &  &  &  \\ \cline{2-10}
\multicolumn{1}{|c|}{}                          & LeNet5     & \multicolumn{1}{c|}{40\%} & \multicolumn{1}{c|}{38\%} & \multicolumn{1}{c|}{42\%} & 0.016 & \multicolumn{1}{c|}{64\%} & \multicolumn{1}{c|}{60\%} & \multicolumn{1}{c|}{68\%} & 0.021 &  &  &  \\ \cline{1-10}
\multicolumn{1}{|c|}{Fashion}            & LeNet4     & \multicolumn{1}{c|}{39\%} & \multicolumn{1}{c|}{36\%} & \multicolumn{1}{c|}{43\%} & 0.019 & \multicolumn{1}{c|}{53\%} & \multicolumn{1}{c|}{52\%} & \multicolumn{1}{c|}{55\%} & 0.008 &  &  &  \\ \cline{1-10}
\multicolumn{1}{|c|}{SVHN}                      & LeNet5     & \multicolumn{1}{c|}{47\%} & \multicolumn{1}{c|}{44\%} & \multicolumn{1}{c|}{49\%} & 0.017 & \multicolumn{1}{c|}{62\%} & \multicolumn{1}{c|}{61\%} & \multicolumn{1}{c|}{64\%} & 0.012 &  &  &  \\ \cline{1-10}
\end{tabular}
\\
\vspace{1em}
Note: The values of average, minimum, and maximum FDRs reported in this table have been rounded up to two decimal digits and are represented as percentages. The standard deviation (std) of FDRs has been rounded up to three decimal digits.
\label{Tab:Stability}
\vspace{-5mm}
\end{table}

\section{Threats to Validity} \label{sec:Threats}

We discuss in this section the different threats to the validity of our study and describe how we mitigated them. 

\noindent{\textbf{Internal threats to validity}} concern the causal relationship between the treatment and the outcome. Since \textit{DeepGD} is black-box and relies on extracting feature matrices to measure the diversity of the selected test input sets, an internal threat to validity might be caused by the poor quality representation of inputs. To mitigate this threat, we have relied on VGG-16, one of the most accurate feature extraction models in the literature. This model has been pre-trained on the very large ImageNet dataset that contains more than 14 million labeled images belonging to 22,000 categories. \Rev{Also, in our prior work~\cite{aghababaeyan2021black}, we assessed the use of the GD metric with the VGG-16 model to measure diversity in input sets. Our results showed that the GD metric, in conjunction with our selected feature extraction model, performed well in measuring actual data diversity across all the studied datasets.}
Moreover, \textit{DeepGD} relies on the specification of a few hyperparameters related to NSGA-II. This also applies to several white-box and black-box test selection approaches that we considered in our study. The configuration of the different hyperparameters in our work may induce additional threats. To mitigate them, we have relied on the NSGA-II hyperparameters recommended in the literature~\cite{briand2006using,cobb1993genetic, messaoudi2018search} except for the mutation rate, which was intentionally set higher and experimentally tuned since we customized the mutation operator to take into account both fitness functions, as described in section~\ref{subsec:mutation}.
We have also considered different configurations of the baselines-related hyperparameters according to their original papers.
A last internal threat to validity would be related to randomness when selecting test input sets with \textit{DeepGD}, ATS, and random selection. We addressed this issue by repeating such selection multiple times while considering different input set sizes and different datasets and models. We also reported on the stability results of \textit{DeepGD}, which indicated that it is stable in this context.

\noindent{\textbf{Construct threats to validity}} concern the relation between the theory and the observations made. A construct threat to validity might be due to inaccuracies in estimating DNN faults since detecting faults in DNNs is not as straightforward as in traditional software. To mitigate this threat, we have relied on a SOTA clustering-based fault estimation approach~\cite{aghababaeyan2021black} that has been thoroughly validated on several models and datasets. In our previous work, we conducted various validation experiments demonstrating that mispredicted inputs within the same cluster are typically mispredicted due to the same fault in the DNN model, whereas inputs from different clusters are mispredicted due to distinct faults (root causes). Additionally, our prior work established that clusters of mispredicted inputs are very different from those in the entire test dataset despite the use of the same feature extraction model (i.e., VGG-16). Such difference further reinforces the validity of our chosen fault estimation approach.
We have reused this publicly available approach to obtain accurate fault estimates. Nonetheless, relying on a such fault estimation approach is still far better than just considering mispredicted inputs that are redundant and due to the same root cause in the DNN model.   


\noindent{\textbf{External threats to validity}} concern the generalizability of our study. We mitigate this threat by considering six different combinations of widely used and architecturally distinct models and datasets. We also considered many testing budgets in our experiments and compared our results with nine SOTA baselines for DNN test selection.

\section{Related Work}
\label{Sec:RW}
In this section, we introduce existing work related to our proposed approach from two aspects: test selection and test diversity in the context of DNN models. \\


\noindent\textbf{Test Selection for DNNs.} Test selection approaches proposed for DNN models can be characterized as black-box or white-box, depending on their access requirements to the internals of the DNN model. 
A few black-box test selection approaches for DNNs have been introduced in the literature. They generally rely on the uncertainty of model classifications. For example, Feng \textit{et al.}~\cite{feng2020deepgini} proposed DeepGini, a test selection approach that prioritizes the selection of inputs with higher \textit{Gini} scores~\cite{feng2020deepgini}. They conjecture that if a DNN model is unsure about a classification and outputs similar probabilities for each class, it is more likely to mispredict the test input. Compared to random and coverage-based selection methods, DeepGini was shown to be more effective in uncovering mispredictions~\cite{feng2020deepgini, weiss2022simple}. Similar to their work, we rely on \textit{Gini} scores as one of the fitness functions in our approach. However, we also consider the diversity of the selected test set and rely on a multi-objective genetic search to guide the search toward finding test inputs with high fault-revealing power. 
Li \textit{et al.}~\cite{li2019boosting} introduced Cross Entropy-based Sampling (CES) and Confidence-based Stratified Sampling (CSS), for black-box DNN test selection. These metrics are used to select a small subset of test inputs that accurately reflect the accuracy of the whole test dataset. They show that compared to random sampling, their approach could achieve the same level of precision with about half of the labeled data. Their goal is clearly different from ours. While our focus is to prioritize the selection of test inputs with high fault-revealing capability, their goal is to minimize test sets. 
Also, though Arrieta~\cite{arrieta2022multi} relied on NSGA-II and uncertainty scores for selecting metamorphic follow-up test cases, our goal with \textit{DeepGD} is also different. We use both diversity and DNN uncertainty to guide the search toward test inputs with high fault-revealing power, for a fixed budget. 

Several white-box test selection approaches have been proposed as well. Such approaches generally rely on coverage~\cite{pei2017deepxplore, Ma2018DeepGaugeMT,gerasimou2020importance} or surprise metrics~\cite{kim2019guiding} to select inputs that will be labeled and used for testing DNN models. For example, Pei \textit{et al.}~\cite{pei2017deepxplore} proposed Neuron Coverage (NC), which measures neurons activation rates in DNN models. Ma \textit{et al.}~\cite{Ma2018DeepGaugeMT} proposed \textit{DeepGauge}, a set of coverage metrics for DNN models that consider neurons activation ranges. Kim \textit{et al.}~\cite{kim2019guiding} proposed surprise coverage metrics which measure how surprising test sets are given the training set. The selection of test inputs for all these approaches is based on maximizing coverage or surprise scores. However, several studies~\cite{ma2021test,chen2020deep,aghababaeyan2021black, li2019structural,harel2020neuron,dong2019there,yang2022revisiting,yan2020correlations}, have shown that these white-box metrics are not always effective for guiding DNNs test selection. For example, Ma \textit{et al.}~\cite{ma2021test} performed an empirical study and compared the effectiveness of white-box metrics (coverage and surprise metrics) with black-box uncertainty in guiding test input selection. Results showed that the former have a weak or no correlation with classification accuracy while the latter had a medium to strong correlation. Uncertainty-based metrics not only outperform coverage-based metrics but also lead to faster improvements in retraining. In our work, as mentioned above, we also consider maximizing the uncertainty score of the test inputs as one of our two fitness objectives. 
 
\noindent\textbf{Diversity in DNN Testing.} Many works have studied diversity-based test selection and generation for traditional software~\cite{hemmati2013achieving, feldt2016test, biagiola2019diversity}. The underlying assumption is that there is a strong correlation between test case diversity and fault-revealing power~\cite{hemmati2013achieving}. Their results confirmed this assumption and showed that diversity-based metrics are effective in revealing faults. Inspired by these encouraging results, researchers devised diversity-based approaches for DNN testing~\cite{zhao2022can,aghababaeyan2021black,gao2022adaptive,zohdinasab2021deephyperion}. Zhao \textit{et al.}~\cite{zhao2022can} conducted an empirical study of SOTA test input selection approaches for DNNs~\cite{chen2020practical, li2019boosting, zhou2020cost} and concluded that they have a negative impact on test diversity and suggest that more research is warranted on designing more effective test selection approaches that guarantee test diversity. 

In a recent study, Gao \textit{et al.}~\cite{gao2022adaptive} proposed the adaptive test selection method (ATS) for DNN models that use the differences between model outputs as a behavior diversity metric. Although ATS aims to cover more diverse faults, its test selection is guided only by model outputs. \textit{DeepGD}, however, considers both the uncertainty of model output probabilities and input features' diversity. Moreover, our results show that \textit{DeepGD} outperforms ATS in test input selection for different combinations of models and datasets and for different test subset sizes.  
In our prior work~\cite{aghababaeyan2021black}, we studied three black-box input diversity metrics for testing DNNs, including geometric diversity~\cite{gong2019diversity}, normalized compression~\cite{cohen2014normalized}, and standard deviation.
We investigated the capacity of these metrics to measure actual diversity in input sets and analyzed their fault-revealing power. Our experiments on image datasets showed that geometric diversity outperforms SOTA white-box coverage criteria in terms of fault detection and computational time. However, in that work, we did not study how the GD metric can be used in practice to guide the selection of test inputs for DNN models.

\Rev{\noindent\textbf{Input Selection for Deep Active Learning.}
Deep Active Learning (DAL) involves the incremental selection of inputs to be labeled and used for (re)training DNN models. Active learning typically starts with a model that has been trained using a small, randomly selected set of inputs. Then, in each (re)training iteration, a subset of unlabeled inputs is selected based on a query strategy, manually labeled, and then used to (re)train the model and improve its performance. In particular, the objective of each iteration is to select the most informative inputs for labeling that enable reducing the labeling cost while improving the accuracy of the model. DAL methods can be categorized into three classes based on the used input selection strategy: uncertainty-based, diversity-based, and combined approaches that leverage both diversity and uncertainty to select inputs for (re)training DNN models effectively~\cite{zhan2022comparative,shui2020deep,ash2019deep}. Zhan \textit{et al.}~\cite{zhan2022comparative} conducted a comparative study including 19 DAL approaches from the three categories. They concluded that the combined DAL approaches such as Wasserstein Adversarial Active Learning (WAAL)~\cite{shui2020deep} and Batch Active learning by Diverse Gradient Embeddings (BADGE)~\cite{ash2019deep}, provide in general better guidance for model retraining compared to uncertainty and diversity-based DAL techniques. Both BADGE and WAAL propose a combined approach for input selection based on uncertainty and diversity.} 

Although BADGE, WAAL, and \textit{DeepGD} appear to share similarities in terms of selection metrics, they diverge in how they measure uncertainty and diversity. For instance, BADGE adopts a white-box approach, relying on the last-layer loss gradients of the model during (re)training to compute uncertainty and construct input feature vectors. It then employs K-means clustering on these feature vectors to select diverse inputs from each cluster. Notably, BADGE primarily operates with DNN models trained using commonly used gradient-based optimizers, unlike \textit{DeepGD} which is agnostic to the model's optimizer.
In addition, WAAL requires the initial labeling of 1\% to 5\% of the (re)training dataset before performing additional and incremental inputs selection. These initially labeled input sets serve as a basis for selecting diverse inputs from the unlabeled dataset, with a focus on those displaying the largest Wasserstein distance w.r.t. the labeled input set~\cite{shui2020deep}.

\Rev{Since active learning techniques require model training-specific data such as gradient loss, uncertainty estimates, or prediction confidence scores during the training epochs~\cite{zhan2022comparative}, they are only applicable during the (re)training phase of DNN models. In contrast, test selection techniques are applicable in both the training and testing phases.
It is also worth noting that active learning techniques require human-in-the-loop labeling throughout the entire (re)training process. This iterative selection of inputs during training iterations makes such approaches impractical in many situations. On the other hand, test selection techniques, including \textit{DeepGD}, only require labeling once, within a given testing budget, before testing and retraining the model.} 
Based on the above points, active learning techniques are therefore not adapted to the objectives of automatically and optimally supporting test selection and using the selected test inputs to support retraining, without requiring iterative human intervention for labeling inputs through the retraining process.

\section{Conclusion}
\label{Sec:Conclusion}

In this paper, we propose \textit{DeepGD}, a multi-objective search-based test input selection approach for DNN models. Our motivation is to provide a black-box test selection mechanism that reduces labeling costs by effectively selecting unlabeled test inputs with high fault-revealing power. We rely on both diversity and uncertainty scores to guide the search toward test inputs that reveal diverse faults in DNNs. We conduct an extensive empirical study on six different subjects and compare \textit{DeepGD} with nine state-of-the-art test selection approaches. Our results show that \textit{DeepGD} statistically and consistently outperforms baseline approaches in terms of its ability to reveal faults in DNN models. Selecting diverse inputs with high uncertainty scores with \textit{DeepGD} not only helps detecting more faults in the DNN model for a given test budget, but also significantly improves the accuracy of the model through retraining with an augmented training set. Our results also indicate that the second-best approach for testing or retraining is not consistent across all models and datasets, further supporting the choice of \textit{DeepGD}. We aim to extend our work by studying the application of diversity and uncertainty metrics for other DNN testing purposes such as test minimization and generation. Finally, we need to investigate alternative ways to estimate faults in DNNs \Rev{and study the relationship between the fault detection rate and the model repair capability of test selection methods.}

\section*{Acknowledgements}

This work was supported by a research grant from General Motors, the Canada Research Chair and Discovery Grant programs of the Natural Sciences and Engineering Research Council of Canada (NSERC), and the Science Foundation Ireland grant agreement No. 957254.



\bibliographystyle{IEEEtran}
\bibliography{main.bib}

\appendix
\section*{Appendix 1} \label{Appendix1}

\begin{table*}[ht]
    \centering
    \small
    \caption{The average number of detected faults in each subject with test subset size $\beta=100$}
    \vspace{-1em}
    \resizebox{\textwidth}{!}{
    \begin{tabular}{|cc|    cc|   ccc|   ccc|   cc|  c c c c  |c|       }
    \hline 
    \multicolumn{2}{|c|}{}  &\multicolumn{10}{c|}{White-box}  &\multicolumn{5}{c|}{Black-box}      \\
    \cline{3-17}   
    \multirow{2}{*}{Data} &\multirow{2}{*}{Model}   
    &\multicolumn{2}{c|}{NC} &\multicolumn{3}{c|}{NBC}  &\multicolumn{3}{c|}{SNAC} 
    &\multirow{2}{*}{LSA}  &\multirow{2}{*}{DSA}   
    &\multirow{2}{*}{RS}   &\multirow{2}{*}{MaxP}  &\multirow{2}{*}{Gini}   &\multirow{2}{*}{ATS}   &\multirow{2}{*}{\textbf{DeepGD}}   \\ 
    &       &0  &0.75      &0  &0.5 &1     &0 &0.5 &1      & &        & & & & &        \\ \hline  \hline
    \multirow{2}{*}{Cifar-10} &12 ConvNet        &13 &21     &13 &12 &12     &12 &11 &13    &31 &35  
     &14 &55 &54 &53       &\textbf{59}    \\ 
                                 &ResNet20      &10  &10      &11 &17 &16     &11 &17 &16    &42 &37
     &11 &52 &56 &50       &\textbf{57}  \\  \cline{1-17}
    \multirow{2}{*}{MNIST}   &LeNet1        &25 &15     &16 &19 &12     &16 &18 &12    &31 &23
     &12 &41 &28 &40       &\textbf{42}   \\ 
                                 &LeNet5        &14 &11     &11 &15 &14     &12 &16 &14    &25 &28
     &11 &30 &29 &30.6       &\textbf{34}    \\  \cline{1-17}
                            Fashion  &LeNet4         &5 &18     &14 &15 &14     &14 &15 &14    &17 &34
     &10 &\textbf{39} &\textbf{39} &32       &\textbf{39}  \\  \cline{1-17}
                            SVHN     &LeNet5        &9 &4     &11 &12 &11     &11 &12 &11  &13 &19
     &11 &43 &43 &44       &\textbf{47}   \\  \cline{1-17}    
         
    \end{tabular}
    }
    
    \label{tab: tabapp1}
\end{table*}

\begin{table*}[ht]
    \centering
    \small
    \caption{The average number of detected faults in each subject with test subset size $\beta=300$}
    \vspace{-1em}
    \resizebox{\textwidth}{!}{
    \begin{tabular}{|cc|    cc|   ccc|   ccc|   cc|  c c c c  |c|       }
    \hline 
    \multicolumn{2}{|c|}{}  &\multicolumn{10}{c|}{White-box}  &\multicolumn{5}{c|}{Black-box}      \\
    \cline{3-17}   
    \multirow{2}{*}{Data} &\multirow{2}{*}{Model}   
    &\multicolumn{2}{c|}{NC} &\multicolumn{3}{c|}{NBC}  &\multicolumn{3}{c|}{SNAC} 
    &\multirow{2}{*}{LSA}  &\multirow{2}{*}{DSA}   
    &\multirow{2}{*}{RS}   &\multirow{2}{*}{MaxP}  &\multirow{2}{*}{Gini}   &\multirow{2}{*}{ATS}   &\multirow{2}{*}{\textbf{DeepGD}}   \\ 
    &       &0  &0.75      &0  &0.5 &1     &0 &0.5 &1      & &        & & & & &        \\ \hline  \hline
    \multirow{2}{*}{Cifar-10} &12 ConvNet        &35 &52     &39 &41 &43     &41 &41 &41    &65 &69  
     &41.1 &101 &99 &93.5       &\textbf{109.6}    \\ 
                                 &ResNet20      &26  &30      &30 &30 &33     &30 &30 &33    &87 &80
     &33.6 &99 &102 &92       &\textbf{104.4}  \\  \cline{1-17}
    \multirow{2}{*}{MNIST}   &LeNet1        &34 &34     &45 &38 &37     &49 &40 &37    &55 &55
     &34.3 &73 &60 &72.6       &\textbf{74}   \\ 
                                 &LeNet5        &29 &28     &27 &28 &25     &27 &25 &25    &46 &44
     &20.4 &54 &50 &52.7       &\textbf{54.4}    \\  \cline{1-17}
                            Fashion  &LeNet4         &11 &35     &27 &32 &32     &27 &32 &32    &49 &63
     &24 &72 &66 &69       &\textbf{74.7}  \\  \cline{1-17}
                            SVHN     &LeNet5        &16 &19     &25 &23 &26     &25 &23 &26  &28 &53
     &25 &85 &90 &85.3       &\textbf{91.1}   \\  \cline{1-17}    
    \end{tabular}
    }
    
    \label{tab: tabapp2}
\end{table*}

\begin{table*}[ht]
    \centering
    
    \small
    \caption{The average number of detected faults in each subject with test subset size $\beta=500$}
    \vspace{-1em}
    \resizebox{\textwidth}{!}{
    \begin{tabular}{|cc|    cc|   ccc|   ccc|   cc|  c c c c  |c|       }
    \hline 
    \multicolumn{2}{|c|}{}  &\multicolumn{10}{c|}{White-box}  &\multicolumn{5}{c|}{Black-box}      \\
    \cline{3-17}   
    \multirow{2}{*}{Data} &\multirow{2}{*}{Model}   
    &\multicolumn{2}{c|}{NC} &\multicolumn{3}{c|}{NBC}  &\multicolumn{3}{c|}{SNAC} 
    &\multirow{2}{*}{LSA}  &\multirow{2}{*}{DSA}   
    &\multirow{2}{*}{RS}   &\multirow{2}{*}{MaxP}  &\multirow{2}{*}{Gini}   &\multirow{2}{*}{ATS}   &\multirow{2}{*}{\textbf{DeepGD}}   \\ 
    &       &0  &0.75      &0  &0.5 &1     &0 &0.5 &1      & &        & & & & &        \\ \hline  \hline
    \multirow{2}{*}{Cifar-10} &12 ConvNet        &56 &64     &60 &62 &56     &56 &58 &54    &84 &95  
     &61.7 &125 &122 &119.8       &\textbf{131.5}    \\ 
                                 &ResNet20      &50  &50      &50 &51 &53     &50 &51 &53    &108 &112
     &47.8 &120 &126 &111.5       &\textbf{129.5}  \\  \cline{1-17}
    \multirow{2}{*}{MNIST}   &LeNet1        &40 &42     &59 &53 &49     &58 &52 &49    &71 &79
     &46.6 &90 &79 &89.1       &\textbf{92}   \\ 
                                 &LeNet5        &33 &36     &35 &39 &34     &36 &36 &34    &55 &55
     &32.3 &62 &61 &60.3       &\textbf{63}    \\  \cline{1-17}
                            Fashion  &LeNet4         &18 &39     &39 &41 &38     &39 &41 &38    &65 &76
     &38 &\textbf{99} &92 &90       &\textbf{99}  \\  \cline{1-17}
                            SVHN     &LeNet5        &21 &32     &35 &32 &37     &35 &32 &37  &40 &68
     &38.2 &113 &112 &105.8       &\textbf{115.1}   \\  \cline{1-17}    
         
    \end{tabular}
    }
    \label{tab: tabapp3}
\end{table*}

\end{document}